\documentclass[sigconf]{acmart}
\usepackage[false]{anonymous-acm}

\AtBeginDocument{%
  \providecommand\BibTeX{{%
    \normalfont B\kern-0.5em{\scshape i\kern-0.25em b}\kern-0.8em\TeX}}}

\setcopyright{acmcopyright}
\copyrightyear{2021}
\acmYear{2021}
\acmDOI{10.1145/nnnnnnn.nnnnnnn}

\acmConference[GECCO '22]{the Genetic and Evolutionary Computation Conference 2022}{July 9--13, 2022}{Boston, USA}
\acmPrice{15.00}
\acmISBN{978-x-xxxx-xxxx-x/YY/MM}

\begin{document}

\title{Solving Multi-Structured Problems by Introducing Linkage Kernels into GOMEA}

\authoranon{
\author{Arthur Guijt}
\email{Arthur.Guijt@cwi.nl}
\orcid{0000-0002-0480-2129}
\affiliation{%
  \institution{Centrum Wiskunde \& Informatica}
  \city{Amsterdam}
  \country{The Netherlands}
}

\author{Dirk Thierens}
\email{D.Thierens@uu.nl}
\affiliation{%
  \institution{Utrecht University}
  \city{Utrecht}
  \country{The Netherlands}
}

\author{Tanja Alderliesten}
\email{T.Alderliesten@lumc.nl}
\orcid{0000-0003-4261-7511}
\affiliation{%
  \institution{Leiden University Medical Center}
  \city{Leiden}
  \country{The Netherlands}
}

\author{Peter A.N. Bosman}
\email{Peter.Bosman@cwi.nl}
\affiliation{%
  \institution{Centrum Wiskunde \& Informatica}
  \city{Amsterdam}
  \country{The Netherlands}
}
\affiliation{%
  \institution{Delft University of Technology}
  \city{Delft}
  \country{The Netherlands}
}
}


\begin{abstract}
  Model-Based Evolutionary Algorithms (MBEAs) can be highly scalable by virtue of linkage (or variable interaction) learning. This requires, however, that the linkage model can capture the exploitable structure of a problem. Usually, a single type of linkage structure is attempted to be captured using models such as a linkage tree. However, in practice, problems may exhibit multiple linkage structures. This is for instance the case in multi-objective optimization when the objectives have different linkage structures. This cannot be modelled sufficiently well when using linkage models that aim at capturing a single type of linkage structure, deteriorating the advantages brought by MBEAs. Therefore, here, we introduce linkage kernels, whereby a linkage structure is learned for each solution over its local neighborhood. We implement linkage kernels into the MBEA known as GOMEA that was previously found to be highly scalable when solving various problems. We further introduce a novel benchmark function called Best-of-Traps (BoT) that has an adjustable degree of different linkage structures. On both BoT and a worst-case scenario-based variant of the well-known MaxCut problem, we experimentally find a vast performance improvement of linkage-kernel GOMEA over GOMEA with a single linkage tree as well as the MBEA known as DSMGA-II.
\end{abstract}



\maketitle

\vspace*{-3mm}
\section{Introduction}

Recombination is most efficient when variables with a strong relationship, i.e., linkage, are recombined jointly~\cite{thierensScalabilityProblemsSimple1999a,whitleyFundamentalPrinciplesDeception1991}. Compared to traditional Evolutionary Algorithms (EAs), Model-based EAs (MBEAs) aim to be significantly more scalable and reliable in solving problems that exhibit a certain type of exploitable problem structure by explicitly modelling aspects of this structure and using this when creating new solutions. Such models may be built based on problem-specific insights, if they are available, or, in case of a black-box optimization scenario, be learned during optimization based on previously performed function evaluations.

In this paper, we consider in particular MBEAs aimed at exploiting information about dependencies between problem variables. Various types of models exist in literature. Models capable of describing complex relationships, such as Bayesian networks~\cite{pelikanBOABayesianOptimization1999}, are however expensive to learn because they capture information about dependencies and explicitly estimate associated probability distributions, making optimization particularly inefficient if function evaluations themselves are not very expensive. Especially in such cases, computationally cheaper alternatives tend to be preferred and have become part of state-of-the-art algorithms in recent years. Examples thereof include the linkage tree
used in the Gene-pool Optimal Mixing Evolutionary Algorithm (GOMEA)~\cite{thierensOptimalMixingEvolutionary2011,dushatskiyParameterlessGenepoolOptimal2021} and the Parameterless Population Pyramid (P3)~\cite{goldmanParameterlessPopulationPyramid2014}, as the incremental linkage set used in Dependency Structure Matrix Genetic Algorithm  (DSMGA-II)~\cite{hsuOptimizationPairwiseLinkage2015}, in which only interactions between variables are modelled explicitly.

Despite many advances in recent years, especially when it comes to benchmark problems, most MBEA literature only consider problems in which a single linkage structure is clearly present. However, in practice, it is well possible a problem exhibits multiple linkage structures (in different parts of the search space). For instance, this can easily occur in Multi-Objective (MO) optimization, i.e., because the different objectives may have different linkage structures. 

In early multi-objective MBEAs, as well as the more recent MO-GOMEA, objective-space clustering is used~\cite{pelikanMultiobjectiveHBOAClustering2005,luongMultiobjectiveGenepoolOptimal2014}. While not originally introduced to capture different linkage structures along the front, the underlying rationale is similar: what is important in one extreme region of the Pareto front, is likely not important in another extreme region. I.e., solutions that maximize one objective may look very different from solutions that maximize another. Employing clustering and restricted mating therefore increases the chances of successful (local) variation. Similarly, clustering also offers a first remedy to multiple linkage structures, especially if linkage is learned in each cluster separately. However, the number of clusters is typically taken to be relatively small, because this has been observed to suffice to achieve effective optimization if solutions differ along the Pareto front, but the linkage information does not, as in the trap-inverse trap problem~\cite{pelikanMultiobjectiveHBOAClustering2005,luongMultiobjectiveGenepoolOptimal2014} . However, if linkage information gradually changes along the front as well, which is to be expected if both objectives have \emph{different} linkage structures, coarse-grained clusters may not suffice. However, this has, to the best of our knowledge, not been studied in detail.

The issue of multiple linkage structures in a single problem is not  restricted to MO optimization. Similar issues can occur in a single-objective setting. For instance, single-objective optimization problems can be (highly) multi-modal, and each mode can have its own linkage structure. Such problems too, have, to the best of our knowledge, never been studied in detail (in the context of MBEAs). That is not to say that multi-modal problems have never been tackled before with EAs. Commonly, niching and mating restrictions are used here. For example, in \cite{chenAdaptiveNichingEDA2010} an estimation-of-distribution algorithm (EDA) is adaptively clustered and niched to optimize both a real-valued and discrete problem, learning multiple models over a population, showing improved performance over a conventional EDA.
Similarly, including niching into CMA-ES was investigated in \cite{preussNichingCMAESNearestbetter2010a}, which reports improvements on some multi-modal functions.
Additionally, in \cite{liuDoubleNichedEvolutionaryAlgorithm2018} the applicability of niching on a real-valued multi-modal multi-objective problem is investigated, showing that to obtain good performance on such problems, diversity in both objective space and parameter space is necessary. Still, partially due to the use of particular benchmark problems, none of these studies address the explicit presence of different variables interactions \emph{per niche} and the additional requirements this may bring to bear on model building in EAs, which is what we consider in this paper.

We name these problems with multiple linkage structures, \emph{multi-structured problems}. In the context of MBEAs, one study considers a closely related phenomenon. Specifically, in~\cite{martinsPairwiseIndependenceIts2016} it is shown that multi-modality can be an issue for use of pairwise linkage, as commonly used in DSMGA-II and GOMEA. To resolve this issue, higher-order linkage was learned directly rather than through pairwise combinations. In practice, this can be significantly more costly than pairwise linkage; the cost of learning higher-order blocks grows exponentially with the order required.

In this work, we propose a more scalable approach that is aimed at solving the more general notion of multi-structured problems, by introducing \emph{linkage kernels}. Rather than learning a single linkage model, every solution learns their own linkage model over their own local neighborhood. In addition, to show the impact that the presence of multiple linkage structures has on both existing MBEAs and our newly proposed linkage kernels, we introduce a novel, scalable benchmark function called Best-of-Traps (BoT) that can be used both single- and multi-objectively. BoT is based on the well-known deceptive trap functions~\cite{debSufficientConditionsDeceptive1994} in which the number of linkage structures and the size of key building blocks are parameterized.

The remainder of this work is organized as follows. In Section~\ref{section:gomea} we describe the recent variant of GOMEA, the approach that LK-GOMEA extends. In Section~\ref{section:lk-gomea}, the extensions that make up LK-GOMEA are described. In Section~\ref{section:problems} the benchmark problems used in this work, Best-of-Traps and MaxCut, are described. Subsection~\ref{section:problems/multiobjective} describes relevant aspects related to MO optimization, as both single- and MO experiments are performed. Section~\ref{section:experiments-singleobjective} contains the experimental setup related to the single-objective experiments, and the corresponding results, in which DSMGA-II, GOMEA and LK-GOMEA are compared. Section~\ref{section:experiments-multiobjective} similarly contains the experimental setup and results for the MO experiments. Finally, we end with the discussion in Section~\ref{section:discussion} and the conclusion in Section~\ref{section:conclusion}.


\section{GOMEA \& MO-GOMEA} \label{section:gomea}
We will focus on MBEAs that employ both characteristics of local search and EDAs. Of particular interest are GOMEA~\cite{dushatskiyParameterlessGenepoolOptimal2021}, DSMGA-II~\cite{hsuOptimizationPairwiseLinkage2015}, and P3~\cite{goldmanParameterlessPopulationPyramid2014}, which have shown to perform well on a wide variety of problems and may be considered state of the art among (discrete) MBEAs.
Given that GOMEA was previously also applied to MO problems, in this work we will extend upon GOMEA. 

\subsection{GOMEA}
GOMEA, originally introduced in~\cite{thierensOptimalMixingEvolutionary2011}, is an MBEA that utilizes a Family Of Subsets (FOS) to describe (strong) dependencies between variables in terms of subsets of variables. These subsets are utilized in a recombination scheme named Gene-pool Optimal Mixing (GOM) which is applied to a copy of each solution in the population. This scheme iterates over the FOS elements in a random order, replacing the values in the solution at the problem variables corresponding to the FOS element with those of a random donor from the population. This change is immediately evaluated. If the change yields a worse fitness, it is reverted, otherwise it is accepted. If a solution has not been improved after a certain number of generations, forced improvements are applied: GOM is performed with the current elitist as donor until a single strict improvement is found. If no such improvement can be found for any FOS element, the solution is replaced with the current elitist solution.

\subsection{MO-GOMEA}
In MO-GOMEA~\cite{luongMultiobjectiveGenepoolOptimal2014}, the multi-objective variant of GOMEA, an elitist archive is added to keep track of the current approximation front. Further, an overlapping variant of k-means-like clustering is applied, based on the distances in the objective space. For each cluster, a linkage tree is learned. Any cluster that has the largest mean value for objective $i$ is considered to be a single-objective cluster. For such a cluster,  single-objective GOM is used with respect to objective $i$. For the remaining clusters, GOM is still used, but a change is regarded an improvement if it is weakly Pareto-dominant over the previous state, or if it can be added to the elitist archive.

\subsection{Enhancements}
Recently, some performance enhancements were proposed for\linebreak GOMEA~\cite{dushatskiyParameterlessGenepoolOptimal2021}. Not all changes are applicable to MO-GOMEA, however. We therefore only employ a subset of the changes here.

Originally, a single linkage tree (LT) is learned from the population in GOMEA, and a separate LT is learned from each cluster in MO-GOMEA. Each LT is learned by first computing Normalized Mutual Information (NMI) between all pairs of variables, as opposed to MI in the original GOMEA version~\cite{thierensOptimalMixingEvolutionary2011}. NMI was observed to lead to slightly better performance on a range of problems. The pairwise NMI is subsequently used in the UPGMA hierarchical clustering algorithm to construct the linkage tree. UPGMA iteratively merges sets of variables, starting from all univariate sets with one variable, until all sets are merged into one set containing all variables. Each node in the resulting clustering tree originally represents a subset of variables in the FOS. As in the latest version of GOMEA~\cite{dushatskiyParameterlessGenepoolOptimal2021}, we employ filtering. This removes all elements for which $\mathit{NMI} \approx 0$ and the subtrees of each element for which $\mathit{NMI} \approx 1$.

Unlike the latest version of GOMEA~\cite{dushatskiyParameterlessGenepoolOptimal2021}, we do not include Local Search (LS). Unlike single-objective search, there are multiple possible acceptation criteria for multi-objective optimization. Finding the best way to perform LS is problem-specific, especially for multi-objective optimization, and is considered outside the scope of this article. Moreover, LS can be added to our newly proposed LK-GOMEA in the same way as to GOMEA.


Further, we use the original Interleaved Multi-start Scheme (IMS) instead of the population pyramid introduced in~\cite{goldmanParameterlessPopulationPyramid2014}, in part to allow comparison with DSMGA-II in the same population sizing setup. The IMS~\cite{dushatskiyParameterlessGenepoolOptimal2021} is an online population sizing scheme, inspired by~\cite{harikParameterlessGeneticAlgorithm1999}, that can also be used to control parameters like the number of clusters in MO-GOMEA~\cite{luongImprovingPerformanceMORVGOMEA2018}. Multiple simultaneous populations of increasing size are used. For each $b$ generations of a smaller population, the next larger population undergoes one generation. This pattern recurses, e.g., the larger population after that undergoes one generation for every $b^2$ generations of the smaller population. Smaller populations with an average fitness lower than that of a larger population are stopped. In GOMEA, commonly $b=4$ is used.

Of special note is donor search, originally introduced in P3~\cite{goldmanParameterlessPopulationPyramid2014}. Instead of using a random donor, a donor is sought for which the variables in the subset do not all have the same value as what is currently in the solution. The effect of this is most notable when the population is close to converging. As we utilize locality in this work, the neighborhoods are much more likely to share variable values, increasing the likelihood of donor search being beneficial.

Decomposition-based methods in multi-objective optimization are known to provide better spread of solutions along the approximation front, and to be better suited for higher-dimensional objective spaces~\cite{zhangMOEAMultiobjectiveEvolutionary2007,liMultiobjectiveOptimizationProblems2009}. Therefore, a second version of MO-GOMEA was introduced that leverages scalarizations by means of which each solution is assigned an improvement direction~\cite{luongImprovingPerformanceMORVGOMEA2018}.
This is especially relevant for our work, since linkage kernels are similarly meant to associate a notion of local linkage relations to each solution that are likely important to exploit in order to achieve improvements. Aligning these notions of locality is likely of high value.


\section{Linkage-Kernel GOMEA} \label{section:lk-gomea}




\subsection{A Problem of 2 Modes}\label{section:kernels/twomodes}
\begin{figure}[htbp]
  \centering
  \includegraphics[width=0.7\columnwidth]{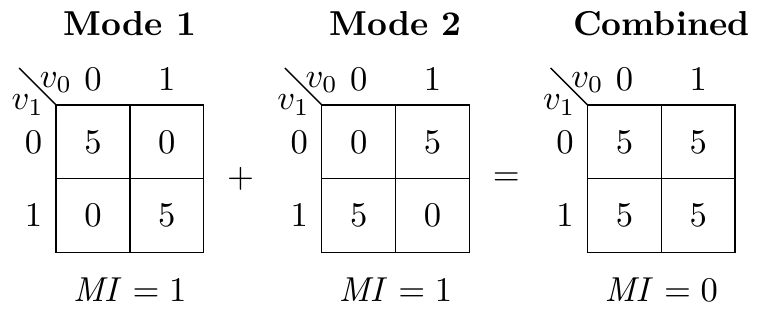}
  \Description{A series of three count matrices for an arbitrary pair of variables and the corresponding inferred mutual information. While the individual subpopulations have MI = 1, MI = 0 for the combination.}
  \caption{Count matrices and corresponding MI values within modes and within the entire search space, for two variables $v_0$ and $v_1$. High MI values for modes individually does not necessarily lead to high MI for the combination.}
  \label{fig:combining-modes}
  \vspace*{-3mm}
\end{figure}

As outlined in the previous Sections, GOMEA (in a black-box optimization setting) employs (normalized) mutual information to detect (pairwise) linkages. The idea is that if two variables are dependent, after selection, some variable value combinations will occur more frequently then others, which can be measured in terms of their mutual information. While this has its limitations, leading to novel dependency detection methods being introduced that do not depend on statistical information~\cite{przewozniczekEmpiricalLinkageLearning2020}, on most single-structure problems, the structure can be efficiently detected. Even so, this changes completely, also for other dependency-detection methods (unless they do a full Walsh decomposition as in~\cite{dushatskiyNovelApproachDesigning2021}), in case of a multi-modal function with different dependencies in each mode. Then, some correlations between variables are undetectable when only looking at a pair of variables. An example of this is given in Figure~\ref{fig:combining-modes}. In this example, we have two modes for which a pair of variables $v_0$ and $v_1$ are strongly correlated, although differently for each mode. Especially at the start of optimization, we can assume the population is equally divided over the modes. That means, that when looking at pairs within the combined population however, the variables will seem to be uncorrelated, i.e. pairwise independence actually occurs. Consequently, models learning linkage from this will not create subsets with these variables together, making it harder to optimize for any mode. The more important the linkage information to optimize each mode, the bigger the problem.

It is important to note that while this example concerns an example of only 2 variables, using higher-order linkage-learning alone would \emph{not} resolve this issue. Firstly, if important linkages are completely different in each mode, but do overlap, the order of linkage learning required could be intractable from a classical single-population statistics point of view. Secondly, the values for high-quality solutions in each mode may very well be very different as well. When recombining during GOM, a subset is essentially sampled from the entire population. Even if the block is perfectly identified for one mode, poor exploitation cannot be avoided without also identifying the locality of the mode and employing restricted mating within a mode. Otherwise, inheriting from solutions from other modes will occur, which will likely still not be effective.

\subsection{Splitting Modes}
Based on the analysis above, it is clear that it is important to adeptly identify modes. If linkage learning works when the modes are separate, splitting up the population and learning separate linkage models for each mode should solve the problem of multiple modes interfering. Luckily, similar to how linkage can be inferred using MI from a population that has experienced selection pressure, solutions belonging to the same mode tend to cluster together. 

In and of itself, this concept is not new to EAs, as it is similar to diversity preserving measures, like mating restrictions and niching, that aim to distribute the population across different modes in the search space different parts of the search space. It has even previously be proposed to learn separate models for separate parts of the search space~\cite{chenAdaptiveNichingEDA2010}.
Similarly related are island topologies used in island model EAs. However, the best topology is known to vary from problem to problem~\cite{rucinskiImpactMigrationTopology2010,sefriouiHierarchicalGeneticAlgorithm2000}.

While these approaches offer the right direction to solve the outlined problem, there is an important catch: how well niching works depends fully on the quality of how the population is split. Including too many solutions from a different mode could have a large negative impact, for the same reasons as outlined in Subsection~\ref{section:kernels/twomodes}. Moreover, unlike the strict boundaries in the approaches above, linkage structures may be shared between different modes, or gradually change along the Pareto front in multi-objective optimization, necessitating many niches or fuzzy clustering boundaries.
Finally, while a solution that has many variable values in common with another solution is likely part of the same mode, a solution with very dissimilar values is not necessarily part of another mode. Splitting the population, especially into a limited number of clusters, is hence a non-trivial problem.

\subsection{Linkage Kernels}
To account for this, we propose to associate a local neighborhood with each solution, from which both solution-specific linkage is learned and donors are sampled from when performing GOM. This enables many different types of linkage structure to be exploited throughout the population, with non-strict neighborhood boundaries, meeting some of the concerns raised above. We refer to this notion as \emph{linkage kernels}.

In this paper, we use 2 definitions of neighborhood. Firstly, we define the neighborhood (or kernel) of a solution as its k-nearest neighbors (KNN) in the population, using Hamming distance as a distance measure, with ties in distance broken randomly. We refer to this neighborhood as the \emph{asymmetric} neighborhood. 

Using KNN as the neighborhood does come with a notable downside. When enough solutions converge to the same point, these solutions can only recombine with one another, and the niche pertaining to the linkage kernel becomes irreversibly converged. We note that KNN is asymmetric: while the converged solutions will only have each other as neighbor, a non-converged individual can still have one of the converged solutions in its neighborhood. Allowing recombination also with these solutions, would still allow for the introduction of new building blocks to the local neighborhood. Therefore, we secondly consider the \emph{symmetric} KNN neighborhood, in which the neighborhood of a solution consists of the solutions of which it is a KNN, in addition to its own KNNs. Note that this means the neighborhood size is no longer exactly $k$ for a solution, but can be larger, depending on the number of neighborhoods that solution occurs in.

What remains is the definition of the size of the neighborhood. This should neither be too big, to avoid including solutions from another mode, or too small, as linkage learning still requires a sufficiently large neighborhood to properly infer linkage.
Partly based on preliminary experiments, we propose to use $k = \left\lceil\sqrt{|P|}\right\rceil$, where $|P|$ is the population size and $k$ is the number of nearest neighbors. This provides a large enough pool to learn local linkage over, yet is small enough to be able to sustain niches. Furthermore, by making $k$ relative to the population size, the IMS will make the size of the neighborhood (note: in terms of number of neighboring solutions in the population) relatively smaller as the population size increases: each doubling of the population size only increases $k$ by a factor of $\sqrt{2}$ (ignoring rounding).

\subsection{LK-GOMEA}
With linkage kernels defined, we can now propose a novel version of GOMEA: Linkage-Kernel GOMEA (LK-GOMEA). In LK-GOMEA, the neighborhoods of each solution are first inferred, either using KNN or its symmetric variant, and then, an LT is learned for each solution based on this neighborhood. When improving a solution through GOM (or FI), the solution's FOS is used instead, and the donor pool is restricted to the neighborhood. Pseudocode can be found in the Supplementary material.

\section{Problems} \label{section:problems}

In this section we provide definitions of the problems we consider.



\subsection{MaxCut \& Worst-of-MaxCuts}
The first problem is weighted MaxCut. In this problem, a graph $G = (V, E)$, with $|V| = \ell$ vertices and weighted edges $e = (v_i, v_j, w_{ij}) \in E$, where $v_i$ and $v_j$ are endpoints and $w_{ij}$ is the corresponding weight, is given.
The goal is to divide the vertices in two sets, such that the sum of weights corresponding to edges between the two sets is maximized.
We encode a solution using a binary string, which determines for each vertex which of the two sets it is in. The objective value given a solution $s$ is then:
\begin{equation}
    f_\text{maxcut}(s) = \sum_{(v_i, v_j, w_{ij}) \in E} \begin{cases}
      w_{ij} &\text{if } s[v_i] \not= s[v_j] \\
        0 & \text{otherwise}
    \end{cases}
\end{equation}
MaxCut is a combinatorial optimization problem that GOMEA is known to solve well~\cite{dushatskiyParameterlessGenepoolOptimal2021}.
It is however a symmetric problem, swapping the sets (i.e., inverting the binary values of a solution) has no impact on the objective value. Semantically similar solutions are therefore not always close in Hamming distance, which could potentially be an issue. Furthermore, in general, weighted MaxCut is an NP-hard problem with many local optima. A single problem instance is used for each string length $\ell$, and consists of a fully connected (dense) graph, with integer weights sampled from $[1, 5]$.

Although weighted MaxCut may have many local optima, the linkage structure of a single instance is still predominantly defined by the graph structure and the weights in a singular way.
We therefore additionally consider a problem we call Worst-of-MaxCuts. Worst-of-MaxCuts is the worst-case scenario-based robust optimization variant of MaxCut. In real-world optimization, oftentimes there are uncertainties and multiple scenarios are then sampled in order to optimize for expected value or to hedge against the worst case. Worst-of-MaxCuts is representative of such problems. In Worst-of-MaxCuts, a set of problem instances is given that all have the same number of vertices. When evaluating a solution, the worst objective value out of the individual MaxCut objective values is returned. In this work, we combine two different instances, generated similarly as above.

\subsection{Best-of-Traps}

While weighted MaxCut and Worst-of-MaxCuts are interesting from a perspective of being a well-known problem and having a relation to real-world problems, to assess the impact of linkage kernels on multi-structured problems in a controlled way, a multi-structured benchmark function is needed. This benchmark function should be scalable, i.e., solvable for increasingly larger sizes, when structure is recognized properly. Furthermore, the degree to which a problem is multi-structured should be adjustable, as to be able to determine to what degree having multiple structures within a problem is an actual issue for scalable optimization.

The concatenated deceptive trap function~\cite{debSufficientConditionsDeceptive1994} is an important problem in the linkage-learning community. Its deceptive nature makes the problem hard to solve in a scalable manner if the underlying structure, i.e., linkage between variables, is not recognized and accounted for~\cite{debSufficientConditionsDeceptive1994,whitleyFundamentalPrinciplesDeception1991}. This makes it a much-used benchmark, used to assess if MBEAs are capable of recognizing higher-order problem structure. Yet, the issues raised around linkage learning have to do with a problem being multi-structured, and this is not the case for the deceptive trap function.


To this end, we here introduce a problem named Best-of-Traps (BoT), defined as the maximum over a number of the sub-problems. Each sub-problem is a \emph{permuted} concatenated deceptive trap function in which we also redefine the unitation function to count not the number of ones, but the number of bits similar to a predefined string, so that the global and local attractor can be made different in the binary space as well (i.e., the optimum is not necessarily all ones for each sub-problem). By taking the maximum, and because each sub-function has the same optimal value, yet different binary encoding of the optimum, multiple modes are introduced. By changing the structure in each mode, multiple linkage structures are introduced that span the entire solution. Concordantly, this problem is multi-modal and has a controllable amount of linkage structure through the number of sub-problems used. This control will allow us to investigate to what extent and from what degree onward, multiple structures may pose an issue.

Each sub-problem in BoT is solvable individually by modern model-based EAs like GOMEA, DMSGA-II and P3. Hence, once a mode is localized, the problem should be effectively solvable. However, since we preserve all modes through the max function, the BoT function effectively contains all linkage structures of all sub-problems, creating situations as illustrated in Figure~\ref{fig:combining-modes}.



For string length $\ell$, each BoT sub-problem consists of $\mathit{fns}$ permuted trap functions $t_a$, $a \in \{0,1,\ldots,\mathit{fns}-1\}$, each with a predefined different optimal solution $s^*_a$, and permutation $\pi_a$, both of length $\ell$. Given block size $k$, each sub-problem is defined to be:

\begin{equation}
    t_a(s) = \sum_{i = 0}^{\ell / k-1} {T\left(\sum_{j = 0}^{k-1} 
    \begin{cases}
        1 & \text{if } s[\pi_a[i k + j]] = s^*_x[\pi_a[i k + j]]\\
        0 & \text{otherwise}
    \end{cases}\right)}
\end{equation}
Where $T$ is the deceptive trap function:
\begin{equation}
    T(u) = \begin{cases}
        k & \text{if } u = k\\
        k - u - 1 & \text{otherwise}
    \end{cases}
\end{equation}
The BoT problem can now be defined as follows:
\begin{equation}
    f_\text{BoT}(s) = \max_{a = 0}^{\mathit{fns} - 1} t_a(s)
\end{equation}

\begin{figure}
    \centering
    \includegraphics[width=0.85\columnwidth]{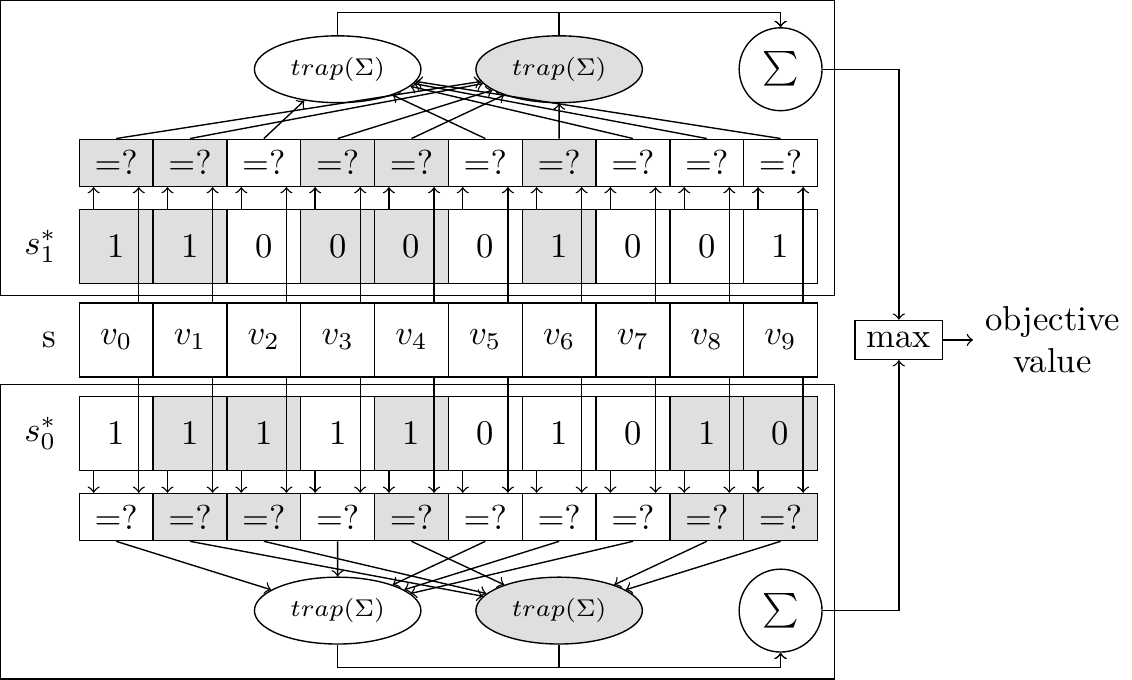}
    \Description{A diagram showing evaluation for an example of a Best-of-Traps instance with 10 variables and two subfunctions. Note that the blocks are different for every subfunction.}
    \caption[Best-of-Traps evaluation example]{Illustration of Best-of-Traps evaluation for $\ell=10$, $k=5$, and $\mathit{fns}=2$. Each sub-problem has their own optimum: $s^*_0$ and $s^*_1$. As $\ell / k = 2$, each sub-problem has two blocks, indicated by white and gray. When a solution $s$ is evaluated, the solution is compared with the optimum for each sub-problem. For each block (color), the number of matches is counted, over which the deceptive trap function is applied. The resulting values are summed together for each sub-problem. The resulting sums are finally combined through a max function.}
    \label{fig:best-of-traps-example}
    \vspace*{-3mm}
\end{figure}

Instances are generated by shuffling indices $(0,1,\ldots,\ell-1)$ to construct $\pi_a$, and uniformly randomly sampling a binary string $s^*_a$ of length $\ell$ for each permuted trap function $t_a$, $a \in \{0,1,\ldots,\mathit{fns}-1\}$. An example with $\mathit{fns} = 2$, $k = 5$ and $\ell = 10$ is given in Figure~\ref{fig:best-of-traps-example}.




\subsection{Multi-Objective Problems} \label{section:problems/multiobjective}
We will furthermore consider multi-objective optimization problems in which the linkage structure gradually changes along the front by virtue of the individual objectives having different linkage structures. Moreover, we will consider situations in which the individual objectives themselves also have multiple linkage structures. Specifically, we propose to perform experiments with BoT vs BoT, BoT vs MaxCut and MaxCut vs MaxCut. For each problem, a different instance is used for each objective.

We will focus especially on the case in which both objectives are BoT. This configuration is interesting due to the combination of both large discrete changes in linkage within one objective, and the gradual change from one objective to another. Of particular interest is how difficult the resulting problem will be, and how well the kernel approach will adapt accordingly.

As these problems are non-trivial to solve, we do not know the optimum (i.e., the Pareto front). To still be able to obtain high-quality fronts for reference, we split up the BoT problem into its sub-problems. We then solved the MO problem for each of pair of sub-problems individually and combined the resulting fronts afterwards. For more details, see the Supplementary material.

\section{Single-objective Experiments} \label{section:experiments-singleobjective}
\subsection{Experimental Setup}
We consider Best-of-Traps (BoT), MaxCut and Worst-of-MaxCuts with problem sizes $\ell \in \{10, 20, 40, 80, 160, 320\}$ for BoT, and $\ell \in \{6, 12, 25, 50, 100, 200\}$ for MaxCut and Worst-of-MaxCuts. We compare LK-GOMEA, using both asymmetric and symmetric linkage kernels, with GOMEA and DSMGA-II~\cite{hsuOptimizationPairwiseLinkage2015}. GOMEA and LK-GOMEA use the LT as linkage structure (per kernel). In order to adapt the population size during a run, all approaches, including DSMGA-II, use the IMS as described in~\cite{dushatskiyParameterlessGenepoolOptimal2021}.

We measure the number of evaluations and number of milliseconds until the optimal solution is found. Limits are set to 100,000,000 evaluations and 6 hours of computational time per run. Each experiment is repeated 30 times. The experiments are performed on 2x AMD EPYC 7282 @ 2.8 GHz with 32 cores total with 252 GB of RAM. At most 30 experiments were running simultaneously, with each experiment being single-threaded.

Our goal is to compare scalability. We therefore perform pairwise statistical tests between GOMEA, DSMGA-II and LK-GOMEA (both variants) for the highest value of $\ell$, and if applicable, for each value of $\mathit{fns}$. Statistical significance tests are performed using the Mann-Whitney U-test~\cite{mannTestWhetherOne1947,fayWilcoxonMannWhitneyTtestAssumptions2010}. Due to the multiplicity of comparisons we perform the Holm-Bonferroni~\cite{holmSimpleSequentiallyRejective1979} correction, with $\alpha=0.05$.

\subsection{Experimental Results \& Discussion}

\begin{figure*}
  \centering
  \includegraphics[width=\textwidth]{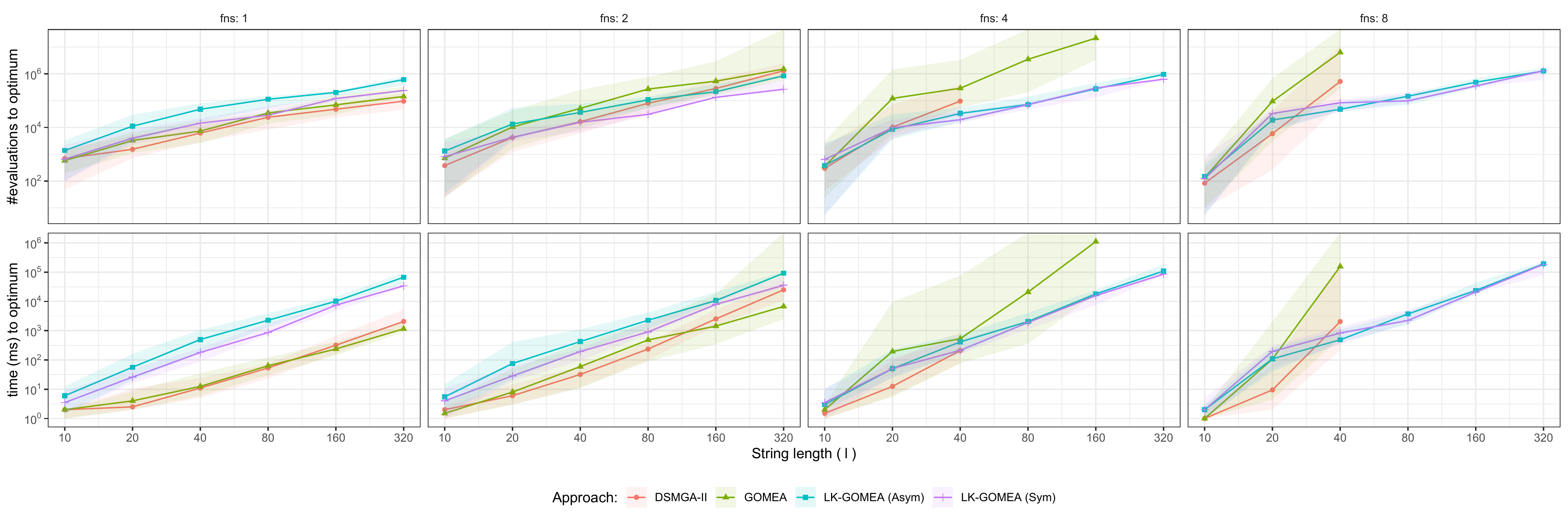}
  \vspace*{-6mm}
  \Description{Scalability graphs displaying the median evaluations and median time required to find the optimum on Best-of-Traps, comparing DSMGA2, GOMEA, and two kernel variants of GOMEA.}
  \caption{Scalability in number of evaluations (top) and time in ms (bottom) on BoT for various problem sizes (i.e. string lengths $\ell$) and number of subfunctions $\mathit{fns}$. The line represents the median, whereas the shaded area ranges from the 95th percentile to the 5th percentile. If the median run was unable to find the optimum within the allotted budget, the point is left out.}
  \label{fig:single-objective-scalability-both-best-of-traps}
\end{figure*}

\begin{figure}
  \centering
  \includegraphics[width=\columnwidth]{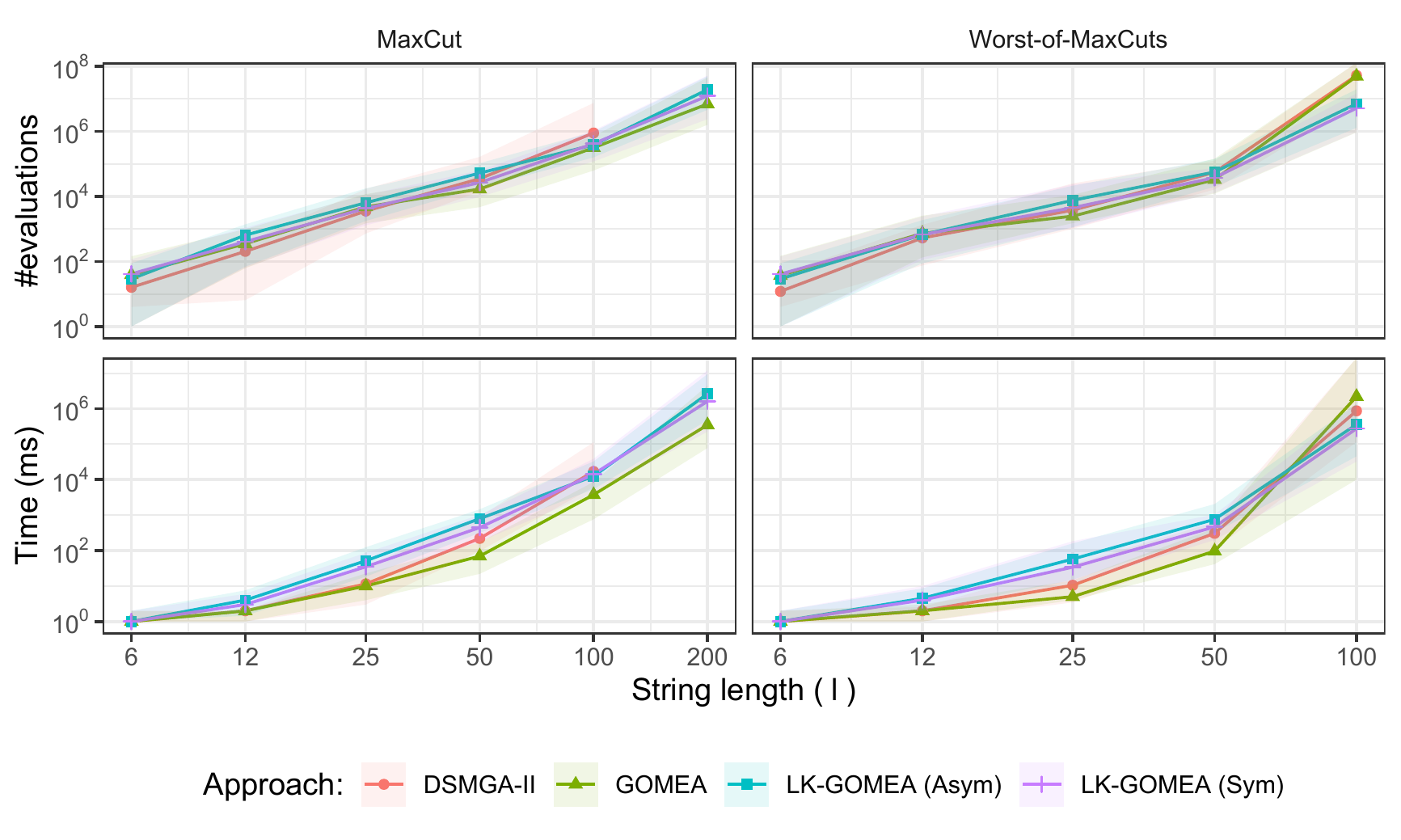}
  \vspace*{-6mm}
  \Description{A scalability graph showing the median number of evaluations required and time required for MaxCut and Worst-of-MaxCuts}
  \caption{Scalability in evaluations to optimum (top) and time to optimum (bottom) on MaxCut (left) and Worst-of-Maxcuts (right) with 2 problem instances, both for various problem sizes (i.e., string lengths). No approach could consistently find the optimum on Worst-of-Maxcuts for $\ell=200$. The line and shaded area are defined similarly to Figure~\ref{fig:single-objective-scalability-both-best-of-traps}.}
  \label{fig:single-objective-scalability-time-and-evaluations-maxcut}
\end{figure}

A summary of the number of settings in which an EA is better (i.e., requires less time or evaluations) is provided in Table~\ref{table:statistical-test-singleobjective}. Details and individual comparisons are in the Supplementary material.

\begin{table}[htbp]
  \caption{Number of times (corresponding rank) an approach is statistically significantly better over all single-objective experiments for number of evaluations-to-optimum (\#eval) and time-to-optimum (time).}
  \label{table:statistical-test-singleobjective}
  \begin{tabular}{cr@{\hspace{1ex}}cc@{\hspace{1ex}}l}
\toprule
       Approach & \multicolumn{2}{c}{\#eval} & \multicolumn{2}{c}{time} \\
\midrule
       DSMGA-II &  3 & (4) &  4 & (3) \\
          GOMEA &  4 & (3) & 10 & (1) \\
LK-GOMEA (Asymmetric) &  9 & (2) &  7 & (2) \\
 LK-GOMEA (Symmetric) & 12 & (1) & 10 & (1) \\
\bottomrule
\end{tabular}

  \vspace*{-3mm}
\end{table}

The scalability results for Best-of-Traps (BoT) can be found in Figure~\ref{fig:single-objective-scalability-both-best-of-traps}. The reliability, i.e., the probability of a run successfully finding the desired optimum within the limits set, can be found in Supplementary material.

When BoT consists of a single sub-problem, BoT is essentially a concatenated deceptive trap function. All approaches can solve BoT in this case scalably and reliably. It is here that the overhead of learning many linkage models in LK-GOMEA is most noticeable, as it is superfluous. While the number of evaluations required increases slightly -- within an order of magnitude -- the amount of time increases by one or two orders of magnitude.

The gap closes as the number of sub-problems in BoT is increased. DSMGA-II fails to solve BoT with four functions (or more) and a problem size larger than 40 variables. Furthermore, at 80 variables and a similar number of sub-problems, GOMEA becomes less reliable. GOMEA is unable to solve problems larger than 40 variables with 8 functions reliably, failing on all runs except for one.

By stark contrast, LK-GOMEA can solve all BoT instances for all problem sizes tested. It is apparent that with eight functions with different linkage, learning multiple linkage models becomes a requirement. LK-GOMEA can solve these problems while maintaining similar scalability as the number of sub-problems grows.


The scalability results for MaxCut and Worst-of-MaxCuts can be found in Figure~\ref{fig:single-objective-scalability-time-and-evaluations-maxcut}. Performance on MaxCut shows similar characteristics as the single-function BoT (i.e., the original permuted deceptive trap function). This is most likely due to the fact that a single MaxCut instance has little variation in linkage structure among its local optima, especially in case of fully connected graphs as utilized here. However, as the problem is more difficult than deceptive trap functions in general, the gap between GOMEA (the best performing EA) and LK-GOMEA is smaller.

For Worst-of-MaxCuts with 2 sub-problems, at $\ell = 100$ LK-GOMEA starts performing significantly better (see Figure~\ref{fig:single-objective-scalability-time-and-evaluations-maxcut}), with a substantial number of runs of GOMEA and DSMGA-II even failing to find the optimum, while LK-GOMEA never failed. This indicates again the usefulness and larger problem-solving capacity of LK-GOMEA also in this worst-case problem. Though not tested, it is expected that the advantage of LK-GOMEA would only increase if the number of MaxCut sub-problems would increase.


\section{Multi-objective Experiments} \label{section:experiments-multiobjective}

\subsection{Experimental Setup}
We again consider Best-of-Traps (BoT) and MaxCut in three configurations: BoT vs BoT, BoT vs MaxCut and MaxCut vs MaxCut. Each objective uses a different instance to ensure that the objectives are not strongly correlated. As the DSMGA-II code used does not support MO optimization, we only use MO-GOMEA~\cite{luongMultiobjectiveGenepoolOptimal2014}. Furthermore, in addition to using a domination-based acceptation criterion, we use the scalarization scheme described in~\cite{luongImprovingPerformanceMORVGOMEA2018}. Finally, for the LK variant, we test both approaches in 2 combinations: using asymmetric (Asym) and symmetric neighborhood (Sym) linkage kernels.

We measure the normalized hypervolume (HV) of the elitist archive over the number of evaluations. HV is the volume/area dominated between a front and a reference point~\cite{nowakEmpiricalPerformanceApproximation2014,zitzlerMultiobjectiveOptimizationUsing1998}. To compute the HV, the ranges of each objective are normalized by the ranges spanned by the reference front. The reference point is taken similarly to \cite{nowakEmpiricalPerformanceApproximation2014}, placing it at an offset of $0.05$ of the worst objective values present, ensuring that the extreme points can contribute HV. The normalized HV is then obtained by dividing the HV of the archive by the HV of the reference front.

Limits on the number of evaluations and time, the hardware used, and other experimental considerations are identical to the single-objective experiments. Statistical tests are also performed similarly as for the single-objective experiments, but now to compare the HV at the evaluation limit for instances where $\ell=100$.

\subsection{Experimental Results \& Discussion}
\begin{figure*}
    \centering
    \includegraphics[width=\textwidth]{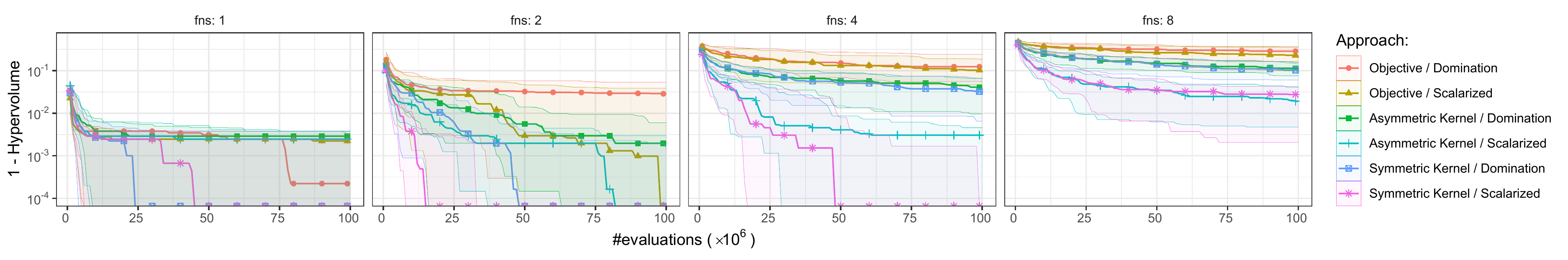}
    \Description{Convergence graphs for Best-of-Traps vs Best-of-Traps for a string length of 100, comparing objective based clustering and kernel based approaches.}
    \caption{Convergence Graphs for $\ell=100$ on BoT (Objective 1) vs BoT (Objective 2), for $\ell=100$ and different numbers of subfunctions (i.e., $fns \in \{1, 2, 4, 8\}$). The points (and the solid line on which they sit) represent the median hypervolume over runs, whereas the shaded area and its bounded area represent the 5th and 95th percentile.}
    \label{fig:multi-objective-convergence-bot-vs-bot-l100}
\end{figure*}
\begin{figure*}
    \centering
    \includegraphics[width=\textwidth]{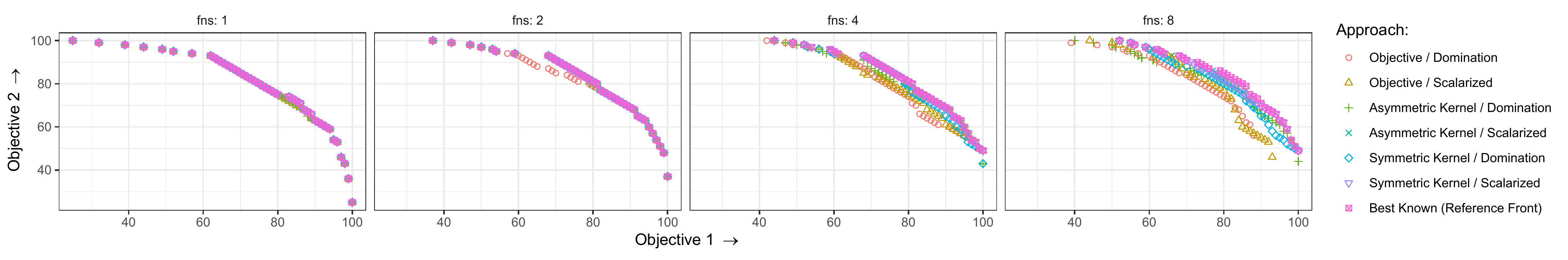}
    \Description{Final fronts obtained by median run for Best-of-Traps vs Best-of-Traps for a string length of 100, comparing objective based clustering and kernel based approaches.}
    \caption{Approximation fronts for the run with the closest-to median obtained hypervolume for $\ell=100$ on BoT (Objective 1) vs BoT (Objective 2).}
    \label{fig:multi-objective-front-bot-vs-bot-l100}
\end{figure*}
\begin{table}
    \caption{Number of times an approach is statistically significantly better on Best-of-Traps vs Best-of-Traps (and the corresponding rank).}
    \label{table:statistical-test-multiobjective}
    \begin{tabular}{lr@{\hspace{1ex}}l}
\toprule
                      Approach & Count & (rank) \\
\midrule
        Objective / Domination &  0 & (6) \\
        Objective / Scalarized &  3 & (5) \\
Asymmetric Kernel / Domination &  5 & (4) \\
Asymmetric Kernel / Scalarized &  9 & (2) \\
 Symmetric Kernel / Domination &  7 & (3) \\
 Symmetric Kernel / Scalarized & 15 & (1) \\
\bottomrule
\end{tabular}

\end{table}

 A summary, stating the number of cases in which an approach is statistically significantly better than another in a setting can be found in Table~\ref{table:statistical-test-multiobjective}. Detailed pairwise tests can be found in Supplementary material.

Convergence graphs for BoT vs BoT, with $\ell=100$ and $\mathit{fns} \in \{1, 2, 4, 8\}$ can be found in Figure~\ref{fig:multi-objective-convergence-bot-vs-bot-l100}. Furthermore, final fronts corresponding to the median run of each approach can be found in Figure~\ref{fig:multi-objective-front-bot-vs-bot-l100}. Graphs and data for BoT vs MaxCut and MaxCut vs MaxCut can be found in the Supplementary material.

The graphs show that objective-space clustering is sufficient when multiple structures along the front are the result of each objective having a single different linkage structure, i.e., when $\mathit{fns} = 1$. However, objective-space clustering becomes increasingly insufficient as $\mathit{fns}$ increases in both objectives. Both variants of LK-GOMEA then perform notably better.

%
Of special note is that BoT seems to be hard to solve around the center part of the front, while the extremes (i.e., single-objective regions) of the front are easier: while the optima for the extremes are found in the median run by LK-GOMEA, in the central region to the front there is a gap to the best solution we found using the pairwise decomposition of the sub-problems as outlined in Section~\ref{section:problems/multiobjective}. Furthermore, although not extensively tested, from several additional runs we did, it appears that solving a weighted combination in single-objective is more difficult in this region, with the MO approach actually obtaining results much closer to the front. A reason for this may be that
at the center of the front, search needs to account for the structure of both objectives. Since in our case, each of the two functions has $\mathit{fns}$ modes, this results in up to $\mathit{fns}^2$ potential modes in the center of the Pareto front. As a multi-objective approach is inherently better at preserving diversity from the perspective of finding multiple solutions along the Pareto front, it is able to maintain and utilize building blocks required to succeed in solving individual functions much better. Additional research is needed however to look into this phenomenon further.

\section{Discussion} \label{section:discussion}
In our work, the neighborhood size has been set to $\sqrt{|P|}$ so as to ensure there was sufficient information to build a linkage model, while being small enough that the locality assumptions would likely hold. The best setting for this parameter is however dependent on the problem as well. A smaller neighborhood is a better fit for problems with a higher degree of multi-modality and multiple linkage structures, whereas a larger neighborhood performs better for problems with no multi-modality. Furthermore, as seen in the MO problem, the number of solutions in each niche can be different. This indicates that a singular value for the neighborhood size is insufficient. Certain modes can vary in their number of solutions, and the neighborhood size needs to be adapted accordingly. An adaptive scheme is therefore of interest for future research.


The fully connected instances for MaxCut do not lend themselves well to linkage learning, and usage of a local neighborhood does not improve this. 
It of interest whether MaxCut always has a more global structure, or if there are instances with a different graph structure or distribution of weights, in which linkage kernels help.

Furthermore, the FI operator performs recombination with the (global) elitist(s). While this is generally beneficial, a solution is not necessarily part of the same mode as the elitist. Applying this operator may therefore waste evaluations and lead to premature replacement of these solutions. An alternative scheme to perform global recombination, or an adaptation of FI that takes into account locality as well, may therefore be of interest for future work.


LK-GOMEA, the new variant of GOMEA evaluated in this work, spends more time on model-building per evaluation than prior versions of GOMEA. The added cost in time may be too great to be applicable on problems where evaluations are computationally very cheap. As the neighborhoods of neighboring individuals tend to be similar, reusing local models or learning linkage models over larger (combined) neighborhoods, could result in a notable improvement without a significant reduction in performance.


Overall, despite the potential possibilities for improvement, linkage kernels have shown to be a useful addition, providing a higher degree of robustness, allowing approaches like LK-GOMEA to solve more complex, multi-structured problems.

\section{Conclusion} \label{section:conclusion}
Model-Based Evolutionary Algorithms can be highly scalable by capturing a problem's structure in a model. In this work we have shown that significantly more useful linkage models can be obtained by introducing linkage kernels, i.e., learning linkage over a local neighborhood, in case of problems that exhibit multiple structures at once. In particular, we have introduced LK-GOMEA, a novel variant of GOMEA that utilizes search space locality by learning a separate linkage model for each solution in the population using a subset of solutions in the population deemed to be in its local neighborhood. While the definition of neighborhood is of special importance, and may well need to be defined differently for particular problems, we observed that a symmetric neighborhood was more robust and led to better performance by LK-GOMEA than using standard (asymmetric) KNN. We have shown LK-GOMEA to be capable of scalably solving a novel benchmark problem that has a tunable number of problem structures where current state-of-the-art model-based evolutionary algorithms such as GOMEA and DSMGA-II fail. Moreover, LK-GOMEA was found to also be superior in solving multi-objective problems that exhibit multiple linkage structures, especially at the problem length and the number of underlying problem structures increases. As this may well occur in complex real-world problems, we believe that we have provided a valuable and novel contribution to the body of work on MBEAs, moving the boundaries of the current state-of-the-art.


\begin{acks}
This publication is part of the project "DAEDALUS - Distributed and Automated Evolutionary Deep Architecture Learning with Unprecedented Scalability" with project number \grantnum{NWO}{18373} of the research programme \grantnum{NWO}{Open Technology Programme} which is (partly) financed by the \grantsponsor{NWO}{Dutch Research Council (NWO)}{}. Other financial contributions as part of this project have been provided by 
\grantsponsor{CElekta}{Elekta AB}{}
 and 
\grantsponsor{CORTECLC}{Ortec Logiqcare B.V.}{}. Special thanks go out to Arkadiy Dushatskiy for providing the source code for DSMGA-II with the Interleaved Multi-start Scheme, as well as the source code of the most recent version of GOMEA from~\cite{dushatskiyParameterlessGenepoolOptimal2021}.
\end{acks}

\bibliographystyle{ACM-Reference-Format}
\bibliography{bibliography.bib}

\end{document}


\maketitle
\tableofcontents
\listoffigures
\listoftables

\section{Reference Solutions \& Fronts}
For MaxCut instances the optima were known and provided. For Best-of-Traps the optima are known a priori from their construction. Fronts for MaxCut vs MaxCut were provided alongside the original MO-GOMEA codebase~\cite{luongMultiobjectiveGenepoolOptimal2018}.

For the single-objective problem Worst-of-Maxcuts the optimum is not known beforehand, and for the new multi-objective problems the full pareto front is not known either. The following subsections discuss and explain how we obtained the fronts for these cases in more detail.

\subsection{Worst-of-Maxcuts}
For Worst-of-Maxcuts no optima are known, instead, a high quality best known value was determined by running GOMEA and other approaches 20 times in parallel, aborting when all runs agreed on the same objective value. From these objective values the best was taken as value-to-reach in order to be deemed 'optimal'.

\subsection{Reference Fronts for Multi-Objective Problems}
Similar to obtaining the reference fronts for Worst-of-Maxcuts, we ran multi-objective GOMEA (with and without linkage kernels) in parallel to obtain reference fronts, with a time limit instead. Unlike the previous approach however, we merged the fronts of all runs together to obtain the reference front.

In order to make the Best-of-Traps problem simpler for the approaches, this problem was decomposed in sub-functions, which is discussed in the next section.

\subsection{Best-of-Traps}
Fronts involving the Best-of-Traps problem were obtained by splitting the Best-of-Traps function into its sub-functions. The resulting sub-problems are then paired with every (sub-)function of the other objective. By running MO-GOMEA (and its variants) for a sufficiently long time \& budget, we obtain fronts for each pair of sub-function between the two objectives. This is done to ensure that the problems are solved without excessive multi-modality, increasing the likelihood of finding the Pareto front to a sub-problem.
Every solution has a chosen (maximal) sub-function for each Best-of-Traps objective. Therefore, any point on the Pareto front for the combined problem, will also be part of the Pareto front of a specific sub-problem.
As such we can obtain the Pareto front for the entire problem by merging the fronts of the sub-problems and removing any dominated points.

\begin{algorithm}
    \caption{LK-GOMEA - GOM, for a full overview of GOMEA, see~\cite{dushatskiyParameterlessGenepoolOptimal2021}}
    \SetKwProg{Fn}{Function}{}{end}
    \SetKwFunction{KwGOM}{GOM} 
    \SetKwFunction{KwFI}{FI} 
    \SetKwFunction{KwOrderFOS}{OrderFOS} 
    \SetKwFunction{KwShuffle}{Shuffle} 
    \SetKwFunction{KwReplace}{Replace} 
    \SetKwFunction{KwEvaluate}{Evaluate} 
    \SetKwFunction{KwAcceptChange}{AcceptChange} 
    \SetKwFunction{KwStrictlyImproved}{StrictlyImproved} 
    \SetKwFunction{KwCopy}{Copy} 
    \SetKw{Break}{break}
    \SetKw{Continue}{continue}

    \Fn{\KwGOM{s, $\mathcal{F}$, \hl{donors}}}{
        \tcc{s is the solution to be improved.
        $\mathcal{F}$ is the FOS.
        donors is now an argument and is -- in case of LK-GOMEA -- the restricted neighborhood of s.}
        \KwOrderFOS{$\mathcal{F}$}\;
        $b \gets \KwCopy{s}$ \tcp*{Make backup}
        $\mathit{changed} \gets \mathit{false}$\;
        \ForEach{$i \in {1, ..., |F|}$}{
            \ForEach{$d \in \KwShuffle{\hl{donors}}$}{
                \KwReplace{s, $F_i$, d} \tcp*{Replace variables corresponding to $F_i$ in s with values from d}
                \If{$b$}
                {
                    \KwEvaluate{s}\;
                    \uIf{\KwAcceptChange{b, s}}{
                        \tcc{For single-objective optimization a change is accepted, if the fitness is equal or better.}
                        $b \gets \KwCopy{s}$\;
                        $\mathit{changed} \gets \mathit{true}$\;
                    }\Else{
                        $s \gets \KwCopy{b}$\;
                    }
                    \Break\;
                }
                \If{$\neg\mathit{use\_donor\_search}$}
                {
                    \Break\;
                }
            }
        }
        \If{$\mathit{use\_fi} \text{ and } (\neg\mathit{changed} \text{ or } \mathit{NIS} > \mathit{NIS\_threshold})$}{
            \tcc{$\mathit{NIS\_threshold}$ is generally $1 + 10 \log_{10}(\mathit{population\_size})$}
            \KwFI{s, $\mathcal{F}$}\;
        }
        \uIf{\KwStrictlyImproved{s}}{
            \tcc{For single-objective optimization a strict improvement means an increase in fitness over its original value.}
            $\mathit{NIS}_s \gets 0$\;
        }\Else{
            $\mathit{NIS}_s \gets \mathit{NIS}_s + 1$\;
        }
    }

\end{algorithm}

\begin{algorithm}
    \caption{LK-GOMEA - Generational Step, for a full overview of GOMEA, see~\cite{dushatskiyParameterlessGenepoolOptimal2021}}
    \SetKwProg{Fn}{Function}{}{end}
    \SetKwFunction{KwStepGOMEA}{StepGOMEA}
    \SetKwFunction{KwGOM}{GOM}
    \SetKwFunction{KwShuffle}{Shuffle}
    \SetKwFunction{KwLearnLT}{LearnLT}
    \SetKwFunction{KwGetNeighborhoods}{GetNeighborhoods}

    \Fn{\KwStepGOMEA{$\mathcal{P}$}}{
        \tcc{$\mathcal{P}$ is the current population.}
        $\mathcal{S} \gets \KwGetNeighborhoods{$\mathcal{P}$}$\;
        $\mathcal{FS} \gets \left\{\KwLearnLT{$S$} \text{ for } S \in \mathcal{S}\right\}$\;
        \ForEach{$i \in \KwShuffle{$\{1, ..., |\mathcal{P}|\}$}$}{
            \KwGOM{$\mathcal{P}_i, \mathcal{FS}_i, \mathcal{S}_i$};
        }
    }
\end{algorithm}

\begin{algorithm}
    \caption{LK-GOMEA - Get Neighborhoods}
    \SetKwProg{Fn}{Function}{}{end}
    \SetKwFunction{KwStepGOMEA}{StepGOMEA} 
    \SetKwFunction{KwShuffle}{Shuffle} 
    \SetKwFunction{KwRandom}{Random} 
    \SetKwFunction{KwPartialSort}{PartialSort} 
    \SetKwFunction{KwDistance}{Distance}
    \SetKwFunction{KwGetNeighborhoods}{GetNeighborhoods}
    \SetKwFunction{KwCopy}{Copy} 
    \SetKw{Break}{break}
    \SetKw{Continue}{continue}

    \Fn{\KwGetNeighborhoods{$\mathcal{P}, k$}}{
        \tcc{$\mathcal{P}$ is the current population, $k$ is the number of nearest neighbors to use.}
        $\mathcal{S} \gets \left\{ \mathcal{S}_i: \emptyset \text{ foreach } i \in \mathcal{P}\right\}$\;
        \ForEach{$i \in \mathcal{P}$}{
            $\mathit{d} \gets \left\{(\KwDistance{i, j}, \KwRandom{}, j) \text{ foreach } j \in \mathcal{P}\right\}$
            \tcc{Reorder such that the $k$-th element is in its sorted position, and the ordering with respect to this element is correct.}
            $\mathit{knn} \gets \KwPartialSort{$d, k$}$\;
            \tcc{Update neighborhood of $i$.}
            $\mathcal{S}_i \gets \mathcal{S}_i \cup \left\{j \text{ foreach } \_, \_, j \in \mathit{knn}[1:k]\right\}$\;
            \tcc{If symmetric KNN neighborhood, add i to the other neighborhoods.}
            \If{$\mathit{symmetric\_kernel}$}{
                \ForEach {$\_, \_, j \in \mathit{knn}[1:k]$}{
                    $\mathcal{S}_j \gets \mathcal{S}_j \cup \{i\}$\;
                }
            }
        }
        \Return S\;
    }
\end{algorithm}

\bibliographystyle{ACM-Reference-Format}
\bibliography{bibliography.bib}

\onecolumn

\begin{figure}
    \centering
    \includegraphics[width=\textwidth]{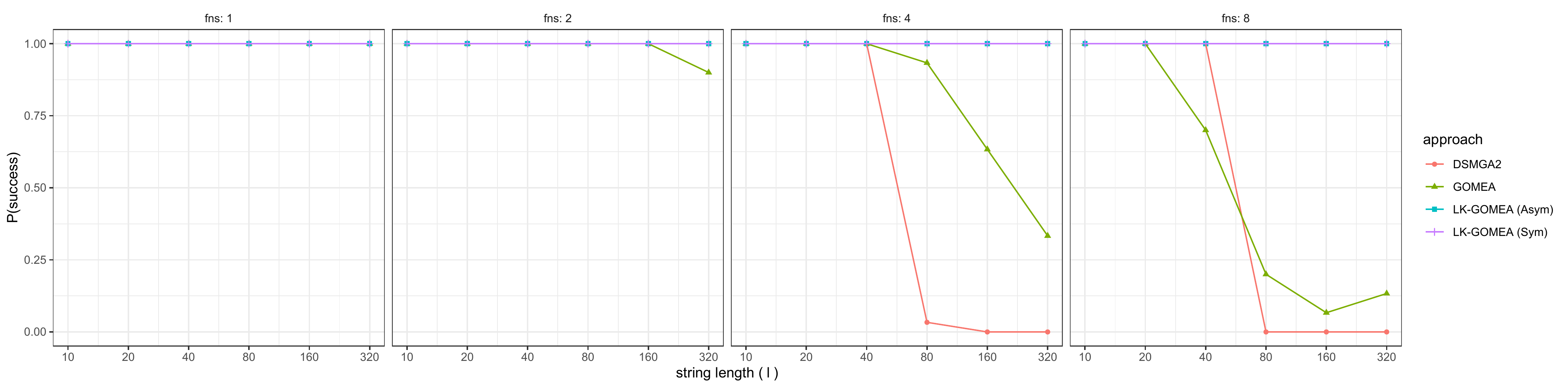}
    \Description[Reliability graphs displaying the ratio of runs that successfully found the optimum for all three approaches.]{ This graph displays the reliability of each approach on Best-of-Traps with varying numbers of sub-functions and string lengths. For a single sub-function all approaches all runs successfully find the optimum. As the number of sub-functions increases behavior of the kernel approaches stays the same. Both GOMEA and DSMGA2 start performing worse with larger numbers of sub-functions, with GOMEA gradually performing worse with larger string lengths as the number of sub-functions increases. DSMGA2 collapses when 4 or more sub-functions are present for a string length of 20 or larger, with a single run finding the optimum for 4 sub-functions and a string length of 20, and no runs successfully finding the optimum otherwise. } 
    \caption{Reliability on Best-of-Traps}
    \label{fig:single-objective-reliability-best-of-traps}
\end{figure}

\begin{figure}
    \centering
    \includegraphics[width=0.5\columnwidth]{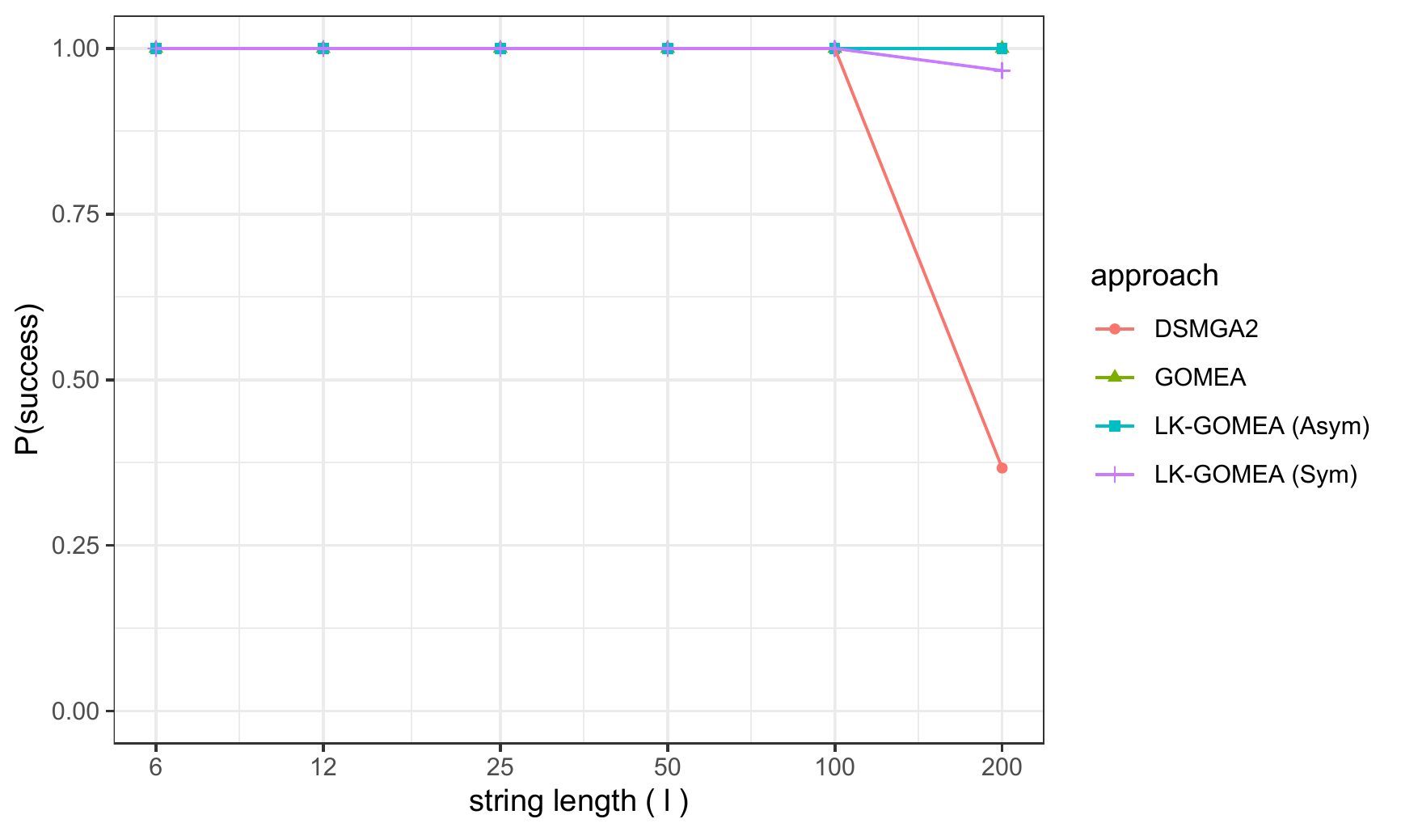}
    \Description[Reliability graphs displaying the ratio of runs that successfully found the optimum for all three approaches.]{ This graph displays the reliability of each approach on MaxCut with varying string lengths. Up to a string length of 100 all approaches and solve the problem reliably. At string length 200 however, DSMGA2 fails to solve the problem reliably. } 
    \caption{Reliability on MaxCut}
    \label{fig:single-objective-reliability-maxcut}
\end{figure}

\begin{figure}
    \centering
    \includegraphics[width=0.5\columnwidth]{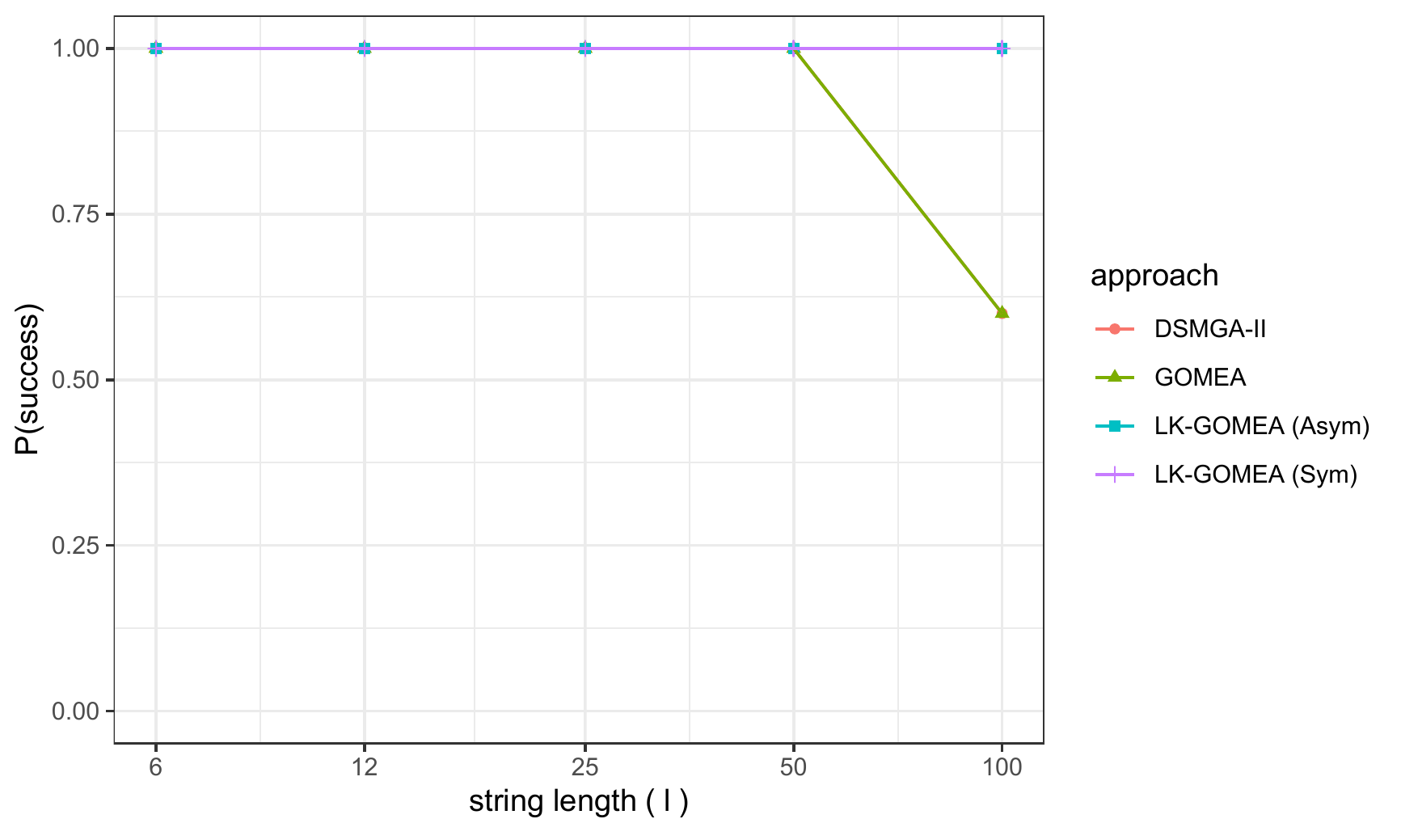}
    \Description[Reliability graphs displaying the ratio of runs that successfully found the optimum for all three approaches.]{ This graph displays the reliability of each approach on MaxCut with varying string lengths. Up to a string length of 50 all approaches and solve the problem reliably. At string length 100 however, DSMGA2 and GOMEA fail to solve the problem in 40\% of the runs. } 
    \caption{Reliability on MaxCut}
    \label{fig:single-objective-reliability-maxcut}
\end{figure}

\begin{table}
    \caption{
        {Medians of \#evaluations to optimum and time (in ms) to optimum for GOMEA, DSMGA-II and LK-GOMEA variants on Best-of-Traps}
    }

\begin{tabular}{rrrrrrrrr}

\toprule
 & \multicolumn{4}{c}{evaluations} & \multicolumn{4}{c}{time (ms)} \\ 
 \cmidrule(lr){2-5} \cmidrule(lr){6-9}
l & {DSMGA2} & {GOMEA} & {LK-GOMEA (Asym)} & {LK-GOMEA (Sym)} & {DSMGA2} & {GOMEA} & {LK-GOMEA (Asym)} & {LK-GOMEA (Sym)} \\ 
\midrule
\multicolumn{1}{l}{fns:  1} \\ 
\midrule
10 & $688$ & $577$ & $1,378$ & $656$ & 2.0 & 2.0 & 6.0 & 3.5 \\ 
20 & $1,529$ & $3,254$ & $11,153$ & $3,956$ & 2.5 & 4.0 & 56.5 & 26.0 \\ 
40 & $6,182$ & $7,330$ & $47,796$ & $14,432$ & 11.0 & 12.5 & 500.0 & 180.5 \\ 
80 & $23,965$ & $34,858$ & $112,771$ & $28,490$ & 53.5 & 63.5 & 2277.5 & 866.5 \\ 
160 & $48,007$ & $69,895$ & $203,282$ & $120,296$ & 319.5 & 238.0 & 10157.0 & 7457.5 \\ 
320 & $95,604$ & $143,387$ & $613,452$ & $238,682$ & 2073.0 & 1160.5 & 67300.0 & 34204.5 \\ 
\midrule
\multicolumn{1}{l}{fns:  2} \\ 
\midrule
10 & $378$ & $710$ & $1,313$ & $816$ & 2.0 & 1.5 & 5.5 & 4.0 \\ 
20 & $4,211$ & $10,402$ & $13,430$ & $4,302$ & 6.0 & 8.0 & 75.5 & 28.5 \\ 
40 & $16,329$ & $51,534$ & $36,330$ & $15,532$ & 32.0 & 59.5 & 425.5 & 193.5 \\ 
80 & $80,021$ & $271,338$ & $107,378$ & $30,702$ & 236.0 & 484.0 & 2281.5 & 911.0 \\ 
160 & $284,547$ & $533,707$ & $215,316$ & $133,144$ & 2530.0 & 1441.0 & 10724.0 & 7952.0 \\ 
320 & $1,304,171$ & $1,519,179$ & $848,955$ & $266,174$ & 24979.5 & 6765.5 & 92261.0 & 35737.0 \\ 
\midrule
\multicolumn{1}{l}{fns:  4} \\ 
\midrule
10 & $296$ & $356$ & $378$ & $641$ & 1.5 & 2.0 & 3.0 & 3.5 \\ 
20 & $10,234$ & $121,485$ & $8,534$ & $9,768$ & 12.5 & 195.5 & 50.5 & 51.0 \\ 
40 & $96,128$ & $290,885$ & $33,371$ & $19,448$ & 209.5 & 533.5 & 415.5 & 214.5 \\ 
80 & NA & $3,510,080$ & $72,342$ & $70,884$ & NA & 20918.5 & 2067.0 & 1912.5 \\ 
160 & NA & $21,626,829$ & $276,122$ & $295,834$ & NA & 1116019.0 & 18101.0 & 15867.5 \\ 
320 & NA & NA & $957,538$ & $632,482$ & NA & NA & 109854.0 & 85707.5 \\ 
\midrule
\multicolumn{1}{l}{fns:  8} \\ 
\midrule
10 & $83$ & $135$ & $146$ & $122$ & 1.0 & 1.0 & 2.0 & 2.0 \\ 
20 & $5,863$ & $94,250$ & $18,931$ & $33,235$ & 9.5 & 106.5 & 109.5 & 196.0 \\ 
40 & $522,099$ & $6,369,578$ & $47,736$ & $83,388$ & 2052.5 & 157626.0 & 489.0 & 831.5 \\ 
80 & NA & NA & $146,551$ & $98,523$ & NA & NA & 3757.5 & 2248.0 \\ 
160 & NA & NA & $488,189$ & $353,224$ & NA & NA & 23853.5 & 21082.0 \\ 
320 & NA & NA & $1,278,739$ & $1,309,482$ & NA & NA & 193228.0 & 185801.0 \\ 
 \bottomrule
\end{tabular}

\end{table}
\begin{table}
    \centering
    \caption{
    {Medians of \#evaluations to optimum and time (in ms) to optimum for GOMEA, DSMGA-II and LK-GOMEA variants on MaxCut}
    }
\begin{tabular}{rrrrrrrrr}
\\ 
\toprule
 & \multicolumn{4}{c}{evaluations} & \multicolumn{4}{c}{time (ms)} \\ 
 \cmidrule(lr){2-5} \cmidrule(lr){6-9}
l & DSMGA2 & GOMEA & LK-GOMEA (Asym) & LK-GOMEA (Sym) & DSMGA2 & GOMEA & LK-GOMEA (Asym) & LK-GOMEA (Sym) \\ 
\midrule
6 & $16$ & $39$ & $29$ & $42$ & 1.0 & 1.0 & 1.0 & 1.0 \\ 
12 & $204$ & $351$ & $644$ & $407$ & 2.0 & 2.0 & 4.0 & 3.0 \\ 
25 & $3,552$ & $4,793$ & $6,390$ & $4,306$ & 11.5 & 10.0 & 51.5 & 35.0 \\ 
50 & $35,211$ & $16,851$ & $52,430$ & $26,734$ & 220.0 & 69.5 & 793.0 & 445.5 \\ 
100 & $895,765$ & $315,905$ & $385,649$ & $420,571$ & 17009.5 & 3758.0 & 12511.5 & 14281.5 \\ 
200 & NA & $6,970,920$ & $19,345,934$ & $12,374,354$ & NA & 348989.0 & 2629480.5 & 1616792.0 \\ 
 \bottomrule
\end{tabular}

\end{table}

\begin{table}
    \centering
    \caption{Medians of \#evaluations to optimum and time (in ms) to optimum for GOMEA, DSMGA-II and LK-GOMEA variants on Worst-of-MaxCuts}
    \begin{tabular}{rrrrrrrrr}
\toprule
 & \multicolumn{4}{c}{evaluations} & \multicolumn{4}{c}{time (ms)} \\ 
 \cmidrule(lr){2-5} \cmidrule(lr){6-9}
l & DSMGA-II & GOMEA & LK-GOMEA (Asym) & LK-GOMEA (Sym) & DSMGA-II & GOMEA & LK-GOMEA (Asym) & LK-GOMEA (Sym) \\ 
\midrule
6 & $12$ & $37$ & $29$ & $41$ & 1.0 & 1.0 & 1.0 & 1.0 \\ 
12 & $530$ & $731$ & $668$ & $673$ & 2.0 & 2.0 & 4.5 & 4.0 \\ 
25 & $3,786$ & $2,478$ & $7,592$ & $4,506$ & 10.5 & 5.0 & 57.5 & 34.0 \\ 
50 & $55,128$ & $32,995$ & $56,463$ & $37,645$ & 306.0 & 96.5 & 759.5 & 465.0 \\ 
100 & $52,927,575$ & $49,709,153$ & $7,095,109$ & $5,209,439$ & 870892.5 & 2203303.0 & 360411.5 & 275196.5 \\ 
 \bottomrule
\end{tabular}
\end{table}

\begin{table}
    \centering
    \caption[Best-of-Traps statistical significance evaluations]{Is x statistically significantly better than y on Best-of-Traps in \#evaluations?}
    \begin{tabular}{llrrrr}
\toprule
{} & {approach y} & {DSMGA2} & {GOMEA} & {LK-GOMEA (Asym)} & {LK-GOMEA (Sym)} \\
{fns} & {approach x} & {} & {} & {} & {} \\
\midrule
\multirow[c]{4}{*}{1} & DSMGA2 & {\cellcolor[HTML]{D3D3D3}} \color[HTML]{000000} - & {\cellcolor[HTML]{FFFFFF}} \color[HTML]{000000} 0.000 & {\cellcolor[HTML]{FFFFFF}} \color[HTML]{000000} 0.000 & {\cellcolor[HTML]{FFFFFF}} \color[HTML]{000000} 0.000 \\
 & GOMEA & {\cellcolor[HTML]{808080}} \color[HTML]{FFFFFF} 1.000 & {\cellcolor[HTML]{D3D3D3}} \color[HTML]{000000} - & {\cellcolor[HTML]{FFFFFF}} \color[HTML]{000000} 0.000 & {\cellcolor[HTML]{FFFFFF}} \color[HTML]{000000} 0.000 \\
 & LK-GOMEA (Asym) & {\cellcolor[HTML]{808080}} \color[HTML]{FFFFFF} 1.000 & {\cellcolor[HTML]{808080}} \color[HTML]{FFFFFF} 1.000 & {\cellcolor[HTML]{D3D3D3}} \color[HTML]{000000} - & {\cellcolor[HTML]{808080}} \color[HTML]{FFFFFF} 1.000 \\
 & LK-GOMEA (Sym) & {\cellcolor[HTML]{808080}} \color[HTML]{FFFFFF} 1.000 & {\cellcolor[HTML]{808080}} \color[HTML]{FFFFFF} 1.000 & {\cellcolor[HTML]{FFFFFF}} \color[HTML]{000000} 0.000 & {\cellcolor[HTML]{D3D3D3}} \color[HTML]{000000} - \\
\midrule
 \multirow[c]{4}{*}{2} & DSMGA2 & {\cellcolor[HTML]{D3D3D3}} \color[HTML]{000000} - & {\cellcolor[HTML]{808080}} \color[HTML]{FFFFFF} 0.150 & {\cellcolor[HTML]{808080}} \color[HTML]{FFFFFF} 1.000 & {\cellcolor[HTML]{808080}} \color[HTML]{FFFFFF} 1.000 \\
 & GOMEA & {\cellcolor[HTML]{808080}} \color[HTML]{FFFFFF} 0.850 & {\cellcolor[HTML]{D3D3D3}} \color[HTML]{000000} - & {\cellcolor[HTML]{808080}} \color[HTML]{FFFFFF} 0.999 & {\cellcolor[HTML]{808080}} \color[HTML]{FFFFFF} 1.000 \\
 & LK-GOMEA (Asym) & {\cellcolor[HTML]{FFFFFF}} \color[HTML]{000000} 0.000 & {\cellcolor[HTML]{FFFFFF}} \color[HTML]{000000} 0.001 & {\cellcolor[HTML]{D3D3D3}} \color[HTML]{000000} - & {\cellcolor[HTML]{808080}} \color[HTML]{FFFFFF} 1.000 \\
 & LK-GOMEA (Sym) & {\cellcolor[HTML]{FFFFFF}} \color[HTML]{000000} 0.000 & {\cellcolor[HTML]{FFFFFF}} \color[HTML]{000000} 0.000 & {\cellcolor[HTML]{FFFFFF}} \color[HTML]{000000} 0.000 & {\cellcolor[HTML]{D3D3D3}} \color[HTML]{000000} - \\
\midrule
 \multirow[c]{4}{*}{4} & DSMGA2 & {\cellcolor[HTML]{D3D3D3}} \color[HTML]{000000} - & {\cellcolor[HTML]{808080}} \color[HTML]{FFFFFF} 1.000 & {\cellcolor[HTML]{808080}} \color[HTML]{FFFFFF} 1.000 & {\cellcolor[HTML]{808080}} \color[HTML]{FFFFFF} 1.000 \\
 & GOMEA & {\cellcolor[HTML]{FFFFFF}} \color[HTML]{000000} 0.000 & {\cellcolor[HTML]{D3D3D3}} \color[HTML]{000000} - & {\cellcolor[HTML]{808080}} \color[HTML]{FFFFFF} 1.000 & {\cellcolor[HTML]{808080}} \color[HTML]{FFFFFF} 1.000 \\
 & LK-GOMEA (Asym) & {\cellcolor[HTML]{FFFFFF}} \color[HTML]{000000} 0.000 & {\cellcolor[HTML]{FFFFFF}} \color[HTML]{000000} 0.000 & {\cellcolor[HTML]{D3D3D3}} \color[HTML]{000000} - & {\cellcolor[HTML]{808080}} \color[HTML]{FFFFFF} 1.000 \\
 & LK-GOMEA (Sym) & {\cellcolor[HTML]{FFFFFF}} \color[HTML]{000000} 0.000 & {\cellcolor[HTML]{FFFFFF}} \color[HTML]{000000} 0.000 & {\cellcolor[HTML]{FFFFFF}} \color[HTML]{000000} 0.000 & {\cellcolor[HTML]{D3D3D3}} \color[HTML]{000000} - \\
\midrule
 \multirow[c]{4}{*}{8} & DSMGA2 & {\cellcolor[HTML]{D3D3D3}} \color[HTML]{000000} - & {\cellcolor[HTML]{808080}} \color[HTML]{FFFFFF} 0.980 & {\cellcolor[HTML]{808080}} \color[HTML]{FFFFFF} 1.000 & {\cellcolor[HTML]{808080}} \color[HTML]{FFFFFF} 1.000 \\
 & GOMEA & {\cellcolor[HTML]{808080}} \color[HTML]{FFFFFF} 0.020 & {\cellcolor[HTML]{D3D3D3}} \color[HTML]{000000} - & {\cellcolor[HTML]{808080}} \color[HTML]{FFFFFF} 1.000 & {\cellcolor[HTML]{808080}} \color[HTML]{FFFFFF} 1.000 \\
 & LK-GOMEA (Asym) & {\cellcolor[HTML]{FFFFFF}} \color[HTML]{000000} 0.000 & {\cellcolor[HTML]{FFFFFF}} \color[HTML]{000000} 0.000 & {\cellcolor[HTML]{D3D3D3}} \color[HTML]{000000} - & {\cellcolor[HTML]{808080}} \color[HTML]{FFFFFF} 0.482 \\
 & LK-GOMEA (Sym) & {\cellcolor[HTML]{FFFFFF}} \color[HTML]{000000} 0.000 & {\cellcolor[HTML]{FFFFFF}} \color[HTML]{000000} 0.000 & {\cellcolor[HTML]{808080}} \color[HTML]{FFFFFF} 0.518 & {\cellcolor[HTML]{D3D3D3}} \color[HTML]{000000} - \\
\bottomrule
\end{tabular}

\end{table}

\begin{table}
    \centering
    \caption[Best-of-Traps statistical significance time]{Is x statistically significantly better than y on Best-of-Traps in time (ms)}
    \begin{tabular}{llrrrr}
\toprule
{} & {approach y} & {DSMGA2} & {GOMEA} & {LK-GOMEA (Asym)} & {LK-GOMEA (Sym)} \\
{fns} & {approach x} & {} & {} & {} & {} \\
\midrule
\multirow[c]{4}{*}{1} & DSMGA2 & {\cellcolor[HTML]{D3D3D3}} \color[HTML]{000000} - & {\cellcolor[HTML]{808080}} \color[HTML]{FFFFFF} 1.000 & {\cellcolor[HTML]{FFFFFF}} \color[HTML]{000000} 0.000 & {\cellcolor[HTML]{FFFFFF}} \color[HTML]{000000} 0.000 \\
 & GOMEA & {\cellcolor[HTML]{FFFFFF}} \color[HTML]{000000} 0.000 & {\cellcolor[HTML]{D3D3D3}} \color[HTML]{000000} - & {\cellcolor[HTML]{FFFFFF}} \color[HTML]{000000} 0.000 & {\cellcolor[HTML]{FFFFFF}} \color[HTML]{000000} 0.000 \\
 & LK-GOMEA (Asym) & {\cellcolor[HTML]{808080}} \color[HTML]{FFFFFF} 1.000 & {\cellcolor[HTML]{808080}} \color[HTML]{FFFFFF} 1.000 & {\cellcolor[HTML]{D3D3D3}} \color[HTML]{000000} - & {\cellcolor[HTML]{808080}} \color[HTML]{FFFFFF} 1.000 \\
 & LK-GOMEA (Sym) & {\cellcolor[HTML]{808080}} \color[HTML]{FFFFFF} 1.000 & {\cellcolor[HTML]{808080}} \color[HTML]{FFFFFF} 1.000 & {\cellcolor[HTML]{FFFFFF}} \color[HTML]{000000} 0.000 & {\cellcolor[HTML]{D3D3D3}} \color[HTML]{000000} - \\
\midrule
 \multirow[c]{4}{*}{2} & DSMGA2 & {\cellcolor[HTML]{D3D3D3}} \color[HTML]{000000} - & {\cellcolor[HTML]{808080}} \color[HTML]{FFFFFF} 0.998 & {\cellcolor[HTML]{FFFFFF}} \color[HTML]{000000} 0.000 & {\cellcolor[HTML]{FFFFFF}} \color[HTML]{000000} 0.000 \\
 & GOMEA & {\cellcolor[HTML]{FFFFFF}} \color[HTML]{000000} 0.002 & {\cellcolor[HTML]{D3D3D3}} \color[HTML]{000000} - & {\cellcolor[HTML]{FFFFFF}} \color[HTML]{000000} 0.000 & {\cellcolor[HTML]{FFFFFF}} \color[HTML]{000000} 0.000 \\
 & LK-GOMEA (Asym) & {\cellcolor[HTML]{808080}} \color[HTML]{FFFFFF} 1.000 & {\cellcolor[HTML]{808080}} \color[HTML]{FFFFFF} 1.000 & {\cellcolor[HTML]{D3D3D3}} \color[HTML]{000000} - & {\cellcolor[HTML]{808080}} \color[HTML]{FFFFFF} 1.000 \\
 & LK-GOMEA (Sym) & {\cellcolor[HTML]{808080}} \color[HTML]{FFFFFF} 1.000 & {\cellcolor[HTML]{808080}} \color[HTML]{FFFFFF} 1.000 & {\cellcolor[HTML]{FFFFFF}} \color[HTML]{000000} 0.000 & {\cellcolor[HTML]{D3D3D3}} \color[HTML]{000000} - \\
\midrule
 \multirow[c]{4}{*}{4} & DSMGA2 & {\cellcolor[HTML]{D3D3D3}} \color[HTML]{000000} - & {\cellcolor[HTML]{808080}} \color[HTML]{FFFFFF} 1.000 & {\cellcolor[HTML]{808080}} \color[HTML]{FFFFFF} 1.000 & {\cellcolor[HTML]{808080}} \color[HTML]{FFFFFF} 1.000 \\
 & GOMEA & {\cellcolor[HTML]{FFFFFF}} \color[HTML]{000000} 0.000 & {\cellcolor[HTML]{D3D3D3}} \color[HTML]{000000} - & {\cellcolor[HTML]{808080}} \color[HTML]{FFFFFF} 1.000 & {\cellcolor[HTML]{808080}} \color[HTML]{FFFFFF} 1.000 \\
 & LK-GOMEA (Asym) & {\cellcolor[HTML]{FFFFFF}} \color[HTML]{000000} 0.000 & {\cellcolor[HTML]{FFFFFF}} \color[HTML]{000000} 0.000 & {\cellcolor[HTML]{D3D3D3}} \color[HTML]{000000} - & {\cellcolor[HTML]{808080}} \color[HTML]{FFFFFF} 1.000 \\
 & LK-GOMEA (Sym) & {\cellcolor[HTML]{FFFFFF}} \color[HTML]{000000} 0.000 & {\cellcolor[HTML]{FFFFFF}} \color[HTML]{000000} 0.000 & {\cellcolor[HTML]{FFFFFF}} \color[HTML]{000000} 0.000 & {\cellcolor[HTML]{D3D3D3}} \color[HTML]{000000} - \\
\midrule
 \multirow[c]{4}{*}{8} & DSMGA2 & {\cellcolor[HTML]{D3D3D3}} \color[HTML]{000000} - & {\cellcolor[HTML]{808080}} \color[HTML]{FFFFFF} 0.980 & {\cellcolor[HTML]{808080}} \color[HTML]{FFFFFF} 1.000 & {\cellcolor[HTML]{808080}} \color[HTML]{FFFFFF} 1.000 \\
 & GOMEA & {\cellcolor[HTML]{808080}} \color[HTML]{FFFFFF} 0.020 & {\cellcolor[HTML]{D3D3D3}} \color[HTML]{000000} - & {\cellcolor[HTML]{808080}} \color[HTML]{FFFFFF} 1.000 & {\cellcolor[HTML]{808080}} \color[HTML]{FFFFFF} 1.000 \\
 & LK-GOMEA (Asym) & {\cellcolor[HTML]{FFFFFF}} \color[HTML]{000000} 0.000 & {\cellcolor[HTML]{FFFFFF}} \color[HTML]{000000} 0.000 & {\cellcolor[HTML]{D3D3D3}} \color[HTML]{000000} - & {\cellcolor[HTML]{808080}} \color[HTML]{FFFFFF} 0.979 \\
 & LK-GOMEA (Sym) & {\cellcolor[HTML]{FFFFFF}} \color[HTML]{000000} 0.000 & {\cellcolor[HTML]{FFFFFF}} \color[HTML]{000000} 0.000 & {\cellcolor[HTML]{808080}} \color[HTML]{FFFFFF} 0.021 & {\cellcolor[HTML]{D3D3D3}} \color[HTML]{000000} - \\
\bottomrule
\end{tabular}

\end{table}

\begin{table}
    \centering
    \caption[MaxCut statistical significance evaluations]{Is x statistically significantly better than y on MaxCut in \#evaluations?}
    \begin{tabular}{lrrrr}
\toprule
{approach y} & {DSMGA2} & {GOMEA} & {LK-GOMEA (Asym)} & {LK-GOMEA (Sym)} \\
{approach x} & {} & {} & {} & {} \\
\midrule
DSMGA2 & {\cellcolor[HTML]{D3D3D3}} \color[HTML]{000000} - & {\cellcolor[HTML]{808080}} \color[HTML]{FFFFFF} 1.000 & {\cellcolor[HTML]{808080}} \color[HTML]{FFFFFF} 1.000 & {\cellcolor[HTML]{808080}} \color[HTML]{FFFFFF} 1.000 \\
GOMEA & {\cellcolor[HTML]{FFFFFF}} \color[HTML]{000000} 0.000 & {\cellcolor[HTML]{D3D3D3}} \color[HTML]{000000} - & {\cellcolor[HTML]{808080}} \color[HTML]{FFFFFF} 0.006 & {\cellcolor[HTML]{808080}} \color[HTML]{FFFFFF} 0.068 \\
LK-GOMEA (Asym) & {\cellcolor[HTML]{FFFFFF}} \color[HTML]{000000} 0.000 & {\cellcolor[HTML]{808080}} \color[HTML]{FFFFFF} 0.994 & {\cellcolor[HTML]{D3D3D3}} \color[HTML]{000000} - & {\cellcolor[HTML]{808080}} \color[HTML]{FFFFFF} 0.723 \\
LK-GOMEA (Sym) & {\cellcolor[HTML]{FFFFFF}} \color[HTML]{000000} 0.000 & {\cellcolor[HTML]{808080}} \color[HTML]{FFFFFF} 0.932 & {\cellcolor[HTML]{808080}} \color[HTML]{FFFFFF} 0.277 & {\cellcolor[HTML]{D3D3D3}} \color[HTML]{000000} - \\
\bottomrule
\end{tabular}

\end{table}

\begin{table}
    \centering
    \caption[MaxCut statistical significance time]{Is x statistically significantly better than y on MaxCut in time (ms)}
    \begin{tabular}{lrrrr}
\toprule
{approach y} & {DSMGA2} & {GOMEA} & {LK-GOMEA (Asym)} & {LK-GOMEA (Sym)} \\
{approach x} & {} & {} & {} & {} \\
\midrule
DSMGA2 & {\cellcolor[HTML]{D3D3D3}} \color[HTML]{000000} - & {\cellcolor[HTML]{808080}} \color[HTML]{FFFFFF} 1.000 & {\cellcolor[HTML]{808080}} \color[HTML]{FFFFFF} 1.000 & {\cellcolor[HTML]{808080}} \color[HTML]{FFFFFF} 1.000 \\
GOMEA & {\cellcolor[HTML]{FFFFFF}} \color[HTML]{000000} 0.000 & {\cellcolor[HTML]{D3D3D3}} \color[HTML]{000000} - & {\cellcolor[HTML]{FFFFFF}} \color[HTML]{000000} 0.000 & {\cellcolor[HTML]{FFFFFF}} \color[HTML]{000000} 0.000 \\
LK-GOMEA (Asym) & {\cellcolor[HTML]{FFFFFF}} \color[HTML]{000000} 0.000 & {\cellcolor[HTML]{808080}} \color[HTML]{FFFFFF} 1.000 & {\cellcolor[HTML]{D3D3D3}} \color[HTML]{000000} - & {\cellcolor[HTML]{808080}} \color[HTML]{FFFFFF} 0.633 \\
LK-GOMEA (Sym) & {\cellcolor[HTML]{FFFFFF}} \color[HTML]{000000} 0.000 & {\cellcolor[HTML]{808080}} \color[HTML]{FFFFFF} 1.000 & {\cellcolor[HTML]{808080}} \color[HTML]{FFFFFF} 0.367 & {\cellcolor[HTML]{D3D3D3}} \color[HTML]{000000} - \\
\bottomrule
\end{tabular}

\end{table}

\begin{table}
    \centering
    \caption[Worst-of-MaxCuts statistical significance evaluations]{Is x statistically significantly better than y on Worst-of-MaxCuts in \#evaluations?}
    \begin{tabular}{lrrrr}
\toprule
{approach y} & {DSMGA2} & {GOMEA} & {LK-GOMEA (Asym)} & {LK-GOMEA (Sym)} \\
{approach x} & {} & {} & {} & {} \\
\midrule
DSMGA2 & {\cellcolor[HTML]{D3D3D3}} \color[HTML]{000000} - & {\cellcolor[HTML]{808080}} \color[HTML]{FFFFFF} 0.688 & {\cellcolor[HTML]{808080}} \color[HTML]{FFFFFF} 1.000 & {\cellcolor[HTML]{808080}} \color[HTML]{FFFFFF} 1.000 \\
GOMEA & {\cellcolor[HTML]{808080}} \color[HTML]{FFFFFF} 0.312 & {\cellcolor[HTML]{D3D3D3}} \color[HTML]{000000} - & {\cellcolor[HTML]{808080}} \color[HTML]{FFFFFF} 1.000 & {\cellcolor[HTML]{808080}} \color[HTML]{FFFFFF} 1.000 \\
LK-GOMEA (Asym) & {\cellcolor[HTML]{FFFFFF}} \color[HTML]{000000} 0.000 & {\cellcolor[HTML]{FFFFFF}} \color[HTML]{000000} 0.000 & {\cellcolor[HTML]{D3D3D3}} \color[HTML]{000000} - & {\cellcolor[HTML]{808080}} \color[HTML]{FFFFFF} 0.733 \\
LK-GOMEA (Sym) & {\cellcolor[HTML]{FFFFFF}} \color[HTML]{000000} 0.000 & {\cellcolor[HTML]{FFFFFF}} \color[HTML]{000000} 0.000 & {\cellcolor[HTML]{808080}} \color[HTML]{FFFFFF} 0.267 & {\cellcolor[HTML]{D3D3D3}} \color[HTML]{000000} - \\
\bottomrule
\end{tabular}

\end{table}

\begin{table}
    \centering
    \caption[Worst-of-MaxCuts statistical significance time]{Is x statistically significantly better than y on Worst-of-MaxCuts in time (ms)}
    \begin{tabular}{lrrrr}
\toprule
{approach y} & {DSMGA2} & {GOMEA} & {LK-GOMEA (Asym)} & {LK-GOMEA (Sym)} \\
{approach x} & {} & {} & {} & {} \\
\midrule
DSMGA2 & {\cellcolor[HTML]{D3D3D3}} \color[HTML]{000000} - & {\cellcolor[HTML]{808080}} \color[HTML]{FFFFFF} 0.561 & {\cellcolor[HTML]{808080}} \color[HTML]{FFFFFF} 0.999 & {\cellcolor[HTML]{808080}} \color[HTML]{FFFFFF} 1.000 \\
GOMEA & {\cellcolor[HTML]{808080}} \color[HTML]{FFFFFF} 0.439 & {\cellcolor[HTML]{D3D3D3}} \color[HTML]{000000} - & {\cellcolor[HTML]{808080}} \color[HTML]{FFFFFF} 0.998 & {\cellcolor[HTML]{808080}} \color[HTML]{FFFFFF} 0.997 \\
LK-GOMEA (Asym) & {\cellcolor[HTML]{FFFFFF}} \color[HTML]{000000} 0.001 & {\cellcolor[HTML]{FFFFFF}} \color[HTML]{000000} 0.002 & {\cellcolor[HTML]{D3D3D3}} \color[HTML]{000000} - & {\cellcolor[HTML]{808080}} \color[HTML]{FFFFFF} 0.671 \\
LK-GOMEA (Sym) & {\cellcolor[HTML]{FFFFFF}} \color[HTML]{000000} 0.000 & {\cellcolor[HTML]{FFFFFF}} \color[HTML]{000000} 0.003 & {\cellcolor[HTML]{808080}} \color[HTML]{FFFFFF} 0.329 & {\cellcolor[HTML]{D3D3D3}} \color[HTML]{000000} - \\
\bottomrule
\end{tabular}

\end{table}

\begin{table}
    \centering
    \caption[Best-of-Traps vs Best-of-Traps statistical significance hypervolume]{Is x statistically significantly better than y on Best-of-Traps vs Best-of-Traps in hypervolume at the budget limit}
    \begin{tabular}{llrrrrrr}
\toprule
{} & {approach y} & {\textbf{(O~/~D)}} & {\textbf{(O~/~S)}} & {\textbf{(AK~/~D)}} & {\textbf{(AK~/~S)}} & {\textbf{(SK~/~D)}} & {\textbf{(SK~/~S)}} \\
{fns} & {approach x} & {} & {} & {} & {} & {} & {} \\
\midrule
\multirow[c]{6}{*}{1} & \textbf{(O~/~D)} & {\cellcolor[HTML]{D3D3D3}} \color[HTML]{000000} - & {\cellcolor[HTML]{808080}} \color[HTML]{FFFFFF} 0.718 & {\cellcolor[HTML]{808080}} \color[HTML]{FFFFFF} 0.120 & {\cellcolor[HTML]{808080}} \color[HTML]{FFFFFF} 0.143 & {\cellcolor[HTML]{808080}} \color[HTML]{FFFFFF} 0.714 & {\cellcolor[HTML]{808080}} \color[HTML]{FFFFFF} 0.986 \\
 & \textbf{(O~/~S)} & {\cellcolor[HTML]{808080}} \color[HTML]{FFFFFF} 0.282 & {\cellcolor[HTML]{D3D3D3}} \color[HTML]{000000} - & {\cellcolor[HTML]{FFFFFF}} \color[HTML]{000000} 0.001 & {\cellcolor[HTML]{808080}} \color[HTML]{FFFFFF} 0.004 & {\cellcolor[HTML]{808080}} \color[HTML]{FFFFFF} 0.418 & {\cellcolor[HTML]{808080}} \color[HTML]{FFFFFF} 0.978 \\
 & \textbf{(AK~/~D)} & {\cellcolor[HTML]{808080}} \color[HTML]{FFFFFF} 0.880 & {\cellcolor[HTML]{808080}} \color[HTML]{FFFFFF} 0.999 & {\cellcolor[HTML]{D3D3D3}} \color[HTML]{000000} - & {\cellcolor[HTML]{808080}} \color[HTML]{FFFFFF} 0.757 & {\cellcolor[HTML]{808080}} \color[HTML]{FFFFFF} 0.970 & {\cellcolor[HTML]{808080}} \color[HTML]{FFFFFF} 1.000 \\
 & \textbf{(AK~/~S)} & {\cellcolor[HTML]{808080}} \color[HTML]{FFFFFF} 0.857 & {\cellcolor[HTML]{808080}} \color[HTML]{FFFFFF} 0.996 & {\cellcolor[HTML]{808080}} \color[HTML]{FFFFFF} 0.243 & {\cellcolor[HTML]{D3D3D3}} \color[HTML]{000000} - & {\cellcolor[HTML]{808080}} \color[HTML]{FFFFFF} 0.945 & {\cellcolor[HTML]{808080}} \color[HTML]{FFFFFF} 1.000 \\
 & \textbf{(SK~/~D)} & {\cellcolor[HTML]{808080}} \color[HTML]{FFFFFF} 0.286 & {\cellcolor[HTML]{808080}} \color[HTML]{FFFFFF} 0.582 & {\cellcolor[HTML]{808080}} \color[HTML]{FFFFFF} 0.030 & {\cellcolor[HTML]{808080}} \color[HTML]{FFFFFF} 0.055 & {\cellcolor[HTML]{D3D3D3}} \color[HTML]{000000} - & {\cellcolor[HTML]{808080}} \color[HTML]{FFFFFF} 0.965 \\
 & \textbf{(SK~/~S)} & {\cellcolor[HTML]{808080}} \color[HTML]{FFFFFF} 0.014 & {\cellcolor[HTML]{808080}} \color[HTML]{FFFFFF} 0.022 & {\cellcolor[HTML]{FFFFFF}} \color[HTML]{000000} 0.000 & {\cellcolor[HTML]{FFFFFF}} \color[HTML]{000000} 0.000 & {\cellcolor[HTML]{808080}} \color[HTML]{FFFFFF} 0.035 & {\cellcolor[HTML]{D3D3D3}} \color[HTML]{000000} - \\
\midrule
 \multirow[c]{6}{*}{2} & \textbf{(O~/~D)} & {\cellcolor[HTML]{D3D3D3}} \color[HTML]{000000} - & {\cellcolor[HTML]{808080}} \color[HTML]{FFFFFF} 1.000 & {\cellcolor[HTML]{808080}} \color[HTML]{FFFFFF} 1.000 & {\cellcolor[HTML]{808080}} \color[HTML]{FFFFFF} 1.000 & {\cellcolor[HTML]{808080}} \color[HTML]{FFFFFF} 1.000 & {\cellcolor[HTML]{808080}} \color[HTML]{FFFFFF} 1.000 \\
 & \textbf{(O~/~S)} & {\cellcolor[HTML]{FFFFFF}} \color[HTML]{000000} 0.000 & {\cellcolor[HTML]{D3D3D3}} \color[HTML]{000000} - & {\cellcolor[HTML]{808080}} \color[HTML]{FFFFFF} 0.273 & {\cellcolor[HTML]{808080}} \color[HTML]{FFFFFF} 0.815 & {\cellcolor[HTML]{808080}} \color[HTML]{FFFFFF} 0.998 & {\cellcolor[HTML]{808080}} \color[HTML]{FFFFFF} 1.000 \\
 & \textbf{(AK~/~D)} & {\cellcolor[HTML]{FFFFFF}} \color[HTML]{000000} 0.000 & {\cellcolor[HTML]{808080}} \color[HTML]{FFFFFF} 0.727 & {\cellcolor[HTML]{D3D3D3}} \color[HTML]{000000} - & {\cellcolor[HTML]{808080}} \color[HTML]{FFFFFF} 0.967 & {\cellcolor[HTML]{808080}} \color[HTML]{FFFFFF} 1.000 & {\cellcolor[HTML]{808080}} \color[HTML]{FFFFFF} 1.000 \\
 & \textbf{(AK~/~S)} & {\cellcolor[HTML]{FFFFFF}} \color[HTML]{000000} 0.000 & {\cellcolor[HTML]{808080}} \color[HTML]{FFFFFF} 0.185 & {\cellcolor[HTML]{808080}} \color[HTML]{FFFFFF} 0.033 & {\cellcolor[HTML]{D3D3D3}} \color[HTML]{000000} - & {\cellcolor[HTML]{808080}} \color[HTML]{FFFFFF} 0.992 & {\cellcolor[HTML]{808080}} \color[HTML]{FFFFFF} 1.000 \\
 & \textbf{(SK~/~D)} & {\cellcolor[HTML]{FFFFFF}} \color[HTML]{000000} 0.000 & {\cellcolor[HTML]{FFFFFF}} \color[HTML]{000000} 0.002 & {\cellcolor[HTML]{FFFFFF}} \color[HTML]{000000} 0.000 & {\cellcolor[HTML]{808080}} \color[HTML]{FFFFFF} 0.008 & {\cellcolor[HTML]{D3D3D3}} \color[HTML]{000000} - & {\cellcolor[HTML]{808080}} \color[HTML]{FFFFFF} 0.980 \\
 & \textbf{(SK~/~S)} & {\cellcolor[HTML]{FFFFFF}} \color[HTML]{000000} 0.000 & {\cellcolor[HTML]{FFFFFF}} \color[HTML]{000000} 0.000 & {\cellcolor[HTML]{FFFFFF}} \color[HTML]{000000} 0.000 & {\cellcolor[HTML]{FFFFFF}} \color[HTML]{000000} 0.000 & {\cellcolor[HTML]{808080}} \color[HTML]{FFFFFF} 0.020 & {\cellcolor[HTML]{D3D3D3}} \color[HTML]{000000} - \\
\midrule
 \multirow[c]{6}{*}{4} & \textbf{(O~/~D)} & {\cellcolor[HTML]{D3D3D3}} \color[HTML]{000000} - & {\cellcolor[HTML]{808080}} \color[HTML]{FFFFFF} 0.950 & {\cellcolor[HTML]{808080}} \color[HTML]{FFFFFF} 1.000 & {\cellcolor[HTML]{808080}} \color[HTML]{FFFFFF} 1.000 & {\cellcolor[HTML]{808080}} \color[HTML]{FFFFFF} 1.000 & {\cellcolor[HTML]{808080}} \color[HTML]{FFFFFF} 1.000 \\
 & \textbf{(O~/~S)} & {\cellcolor[HTML]{808080}} \color[HTML]{FFFFFF} 0.050 & {\cellcolor[HTML]{D3D3D3}} \color[HTML]{000000} - & {\cellcolor[HTML]{808080}} \color[HTML]{FFFFFF} 1.000 & {\cellcolor[HTML]{808080}} \color[HTML]{FFFFFF} 1.000 & {\cellcolor[HTML]{808080}} \color[HTML]{FFFFFF} 1.000 & {\cellcolor[HTML]{808080}} \color[HTML]{FFFFFF} 1.000 \\
 & \textbf{(AK~/~D)} & {\cellcolor[HTML]{FFFFFF}} \color[HTML]{000000} 0.000 & {\cellcolor[HTML]{FFFFFF}} \color[HTML]{000000} 0.000 & {\cellcolor[HTML]{D3D3D3}} \color[HTML]{000000} - & {\cellcolor[HTML]{808080}} \color[HTML]{FFFFFF} 1.000 & {\cellcolor[HTML]{808080}} \color[HTML]{FFFFFF} 0.812 & {\cellcolor[HTML]{808080}} \color[HTML]{FFFFFF} 1.000 \\
 & \textbf{(AK~/~S)} & {\cellcolor[HTML]{FFFFFF}} \color[HTML]{000000} 0.000 & {\cellcolor[HTML]{FFFFFF}} \color[HTML]{000000} 0.000 & {\cellcolor[HTML]{FFFFFF}} \color[HTML]{000000} 0.000 & {\cellcolor[HTML]{D3D3D3}} \color[HTML]{000000} - & {\cellcolor[HTML]{FFFFFF}} \color[HTML]{000000} 0.000 & {\cellcolor[HTML]{808080}} \color[HTML]{FFFFFF} 1.000 \\
 & \textbf{(SK~/~D)} & {\cellcolor[HTML]{FFFFFF}} \color[HTML]{000000} 0.000 & {\cellcolor[HTML]{FFFFFF}} \color[HTML]{000000} 0.000 & {\cellcolor[HTML]{808080}} \color[HTML]{FFFFFF} 0.188 & {\cellcolor[HTML]{808080}} \color[HTML]{FFFFFF} 1.000 & {\cellcolor[HTML]{D3D3D3}} \color[HTML]{000000} - & {\cellcolor[HTML]{808080}} \color[HTML]{FFFFFF} 1.000 \\
 & \textbf{(SK~/~S)} & {\cellcolor[HTML]{FFFFFF}} \color[HTML]{000000} 0.000 & {\cellcolor[HTML]{FFFFFF}} \color[HTML]{000000} 0.000 & {\cellcolor[HTML]{FFFFFF}} \color[HTML]{000000} 0.000 & {\cellcolor[HTML]{FFFFFF}} \color[HTML]{000000} 0.000 & {\cellcolor[HTML]{FFFFFF}} \color[HTML]{000000} 0.000 & {\cellcolor[HTML]{D3D3D3}} \color[HTML]{000000} - \\
\midrule
 \multirow[c]{6}{*}{8} & \textbf{(O~/~D)} & {\cellcolor[HTML]{D3D3D3}} \color[HTML]{000000} - & {\cellcolor[HTML]{808080}} \color[HTML]{FFFFFF} 0.996 & {\cellcolor[HTML]{808080}} \color[HTML]{FFFFFF} 1.000 & {\cellcolor[HTML]{808080}} \color[HTML]{FFFFFF} 1.000 & {\cellcolor[HTML]{808080}} \color[HTML]{FFFFFF} 1.000 & {\cellcolor[HTML]{808080}} \color[HTML]{FFFFFF} 1.000 \\
 & \textbf{(O~/~S)} & {\cellcolor[HTML]{808080}} \color[HTML]{FFFFFF} 0.004 & {\cellcolor[HTML]{D3D3D3}} \color[HTML]{000000} - & {\cellcolor[HTML]{808080}} \color[HTML]{FFFFFF} 1.000 & {\cellcolor[HTML]{808080}} \color[HTML]{FFFFFF} 1.000 & {\cellcolor[HTML]{808080}} \color[HTML]{FFFFFF} 1.000 & {\cellcolor[HTML]{808080}} \color[HTML]{FFFFFF} 1.000 \\
 & \textbf{(AK~/~D)} & {\cellcolor[HTML]{FFFFFF}} \color[HTML]{000000} 0.000 & {\cellcolor[HTML]{FFFFFF}} \color[HTML]{000000} 0.000 & {\cellcolor[HTML]{D3D3D3}} \color[HTML]{000000} - & {\cellcolor[HTML]{808080}} \color[HTML]{FFFFFF} 1.000 & {\cellcolor[HTML]{808080}} \color[HTML]{FFFFFF} 0.879 & {\cellcolor[HTML]{808080}} \color[HTML]{FFFFFF} 1.000 \\
 & \textbf{(AK~/~S)} & {\cellcolor[HTML]{FFFFFF}} \color[HTML]{000000} 0.000 & {\cellcolor[HTML]{FFFFFF}} \color[HTML]{000000} 0.000 & {\cellcolor[HTML]{FFFFFF}} \color[HTML]{000000} 0.000 & {\cellcolor[HTML]{D3D3D3}} \color[HTML]{000000} - & {\cellcolor[HTML]{FFFFFF}} \color[HTML]{000000} 0.000 & {\cellcolor[HTML]{808080}} \color[HTML]{FFFFFF} 0.304 \\
 & \textbf{(SK~/~D)} & {\cellcolor[HTML]{FFFFFF}} \color[HTML]{000000} 0.000 & {\cellcolor[HTML]{FFFFFF}} \color[HTML]{000000} 0.000 & {\cellcolor[HTML]{808080}} \color[HTML]{FFFFFF} 0.121 & {\cellcolor[HTML]{808080}} \color[HTML]{FFFFFF} 1.000 & {\cellcolor[HTML]{D3D3D3}} \color[HTML]{000000} - & {\cellcolor[HTML]{808080}} \color[HTML]{FFFFFF} 1.000 \\
 & \textbf{(SK~/~S)} & {\cellcolor[HTML]{FFFFFF}} \color[HTML]{000000} 0.000 & {\cellcolor[HTML]{FFFFFF}} \color[HTML]{000000} 0.000 & {\cellcolor[HTML]{FFFFFF}} \color[HTML]{000000} 0.000 & {\cellcolor[HTML]{808080}} \color[HTML]{FFFFFF} 0.696 & {\cellcolor[HTML]{FFFFFF}} \color[HTML]{000000} 0.000 & {\cellcolor[HTML]{D3D3D3}} \color[HTML]{000000} - \\
\bottomrule
\end{tabular}

\textbf{(O~/~D):}~Objective~/~Domination \hspace{2em}
\textbf{(O~/~S):}~Objective~/~Scalarized \hspace{2em}
\textbf{(AK~/~D):}~Asymmetric Kernel~/~Domination \hspace{2em}
\textbf{(AK~/~S):}~Asymmetric Kernel~/~Scalarized \hspace{2em}
\textbf{(SK~/~D):}~Symmetric Kernel~/~Domination \hspace{2em}
\textbf{(SK~/~S):}~Symmetric Kernel~/~Scalarized

\end{table}

\begin{figure}
    \centering
    \begin{subfigure}{\textwidth}
        \includegraphics[width=\textwidth]{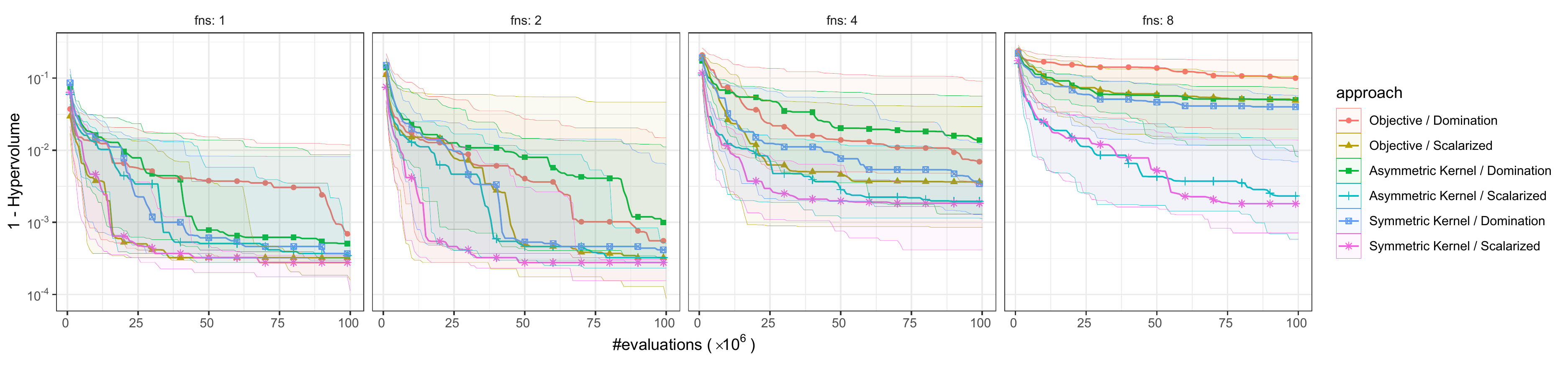}
        \subcaption{Convergence graph}
    \end{subfigure}
    \begin{subfigure}{\textwidth}
        \includegraphics[width=\textwidth]{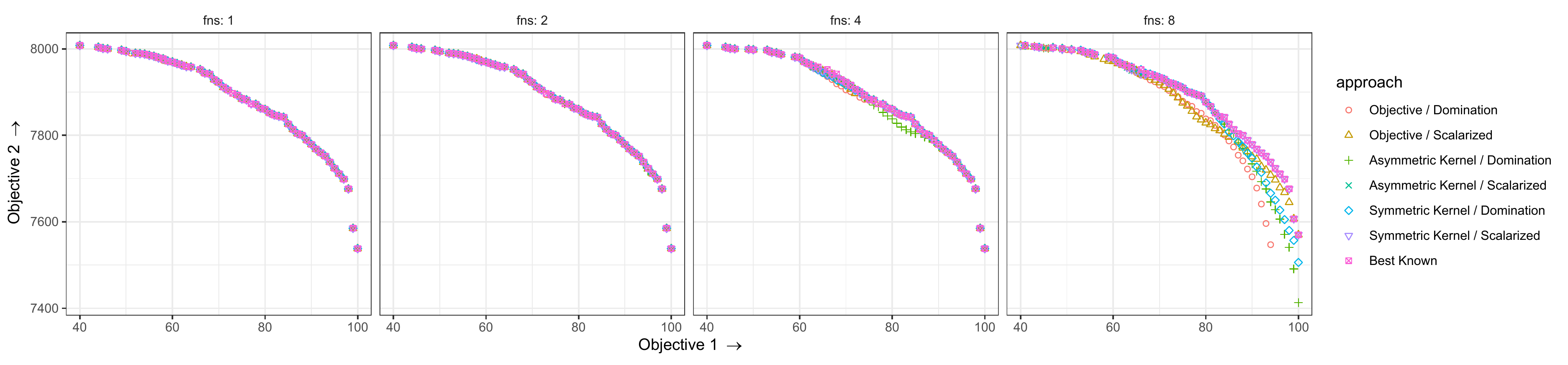}
        \subcaption{Final fronts}
    \end{subfigure}

    \caption{Best-of-Traps vs MaxCut $l=100$}
\end{figure}
\begin{figure}
    \centering
    \begin{subfigure}{0.4\textwidth}
        \includegraphics[width=\textwidth]{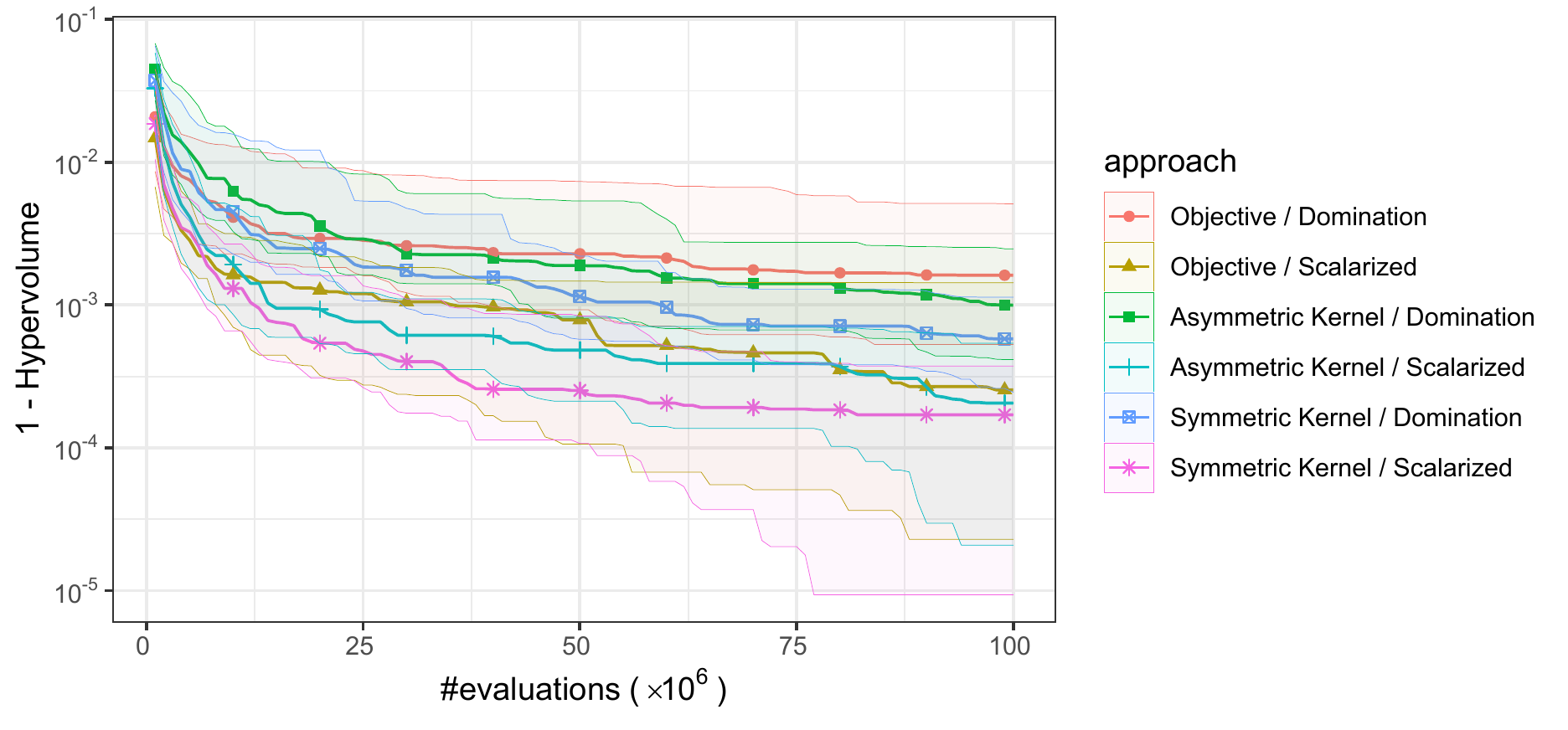}
        \subcaption{Convergence Graph}
    \end{subfigure}
    \begin{subfigure}{0.4\textwidth}
        \includegraphics[width=\textwidth]{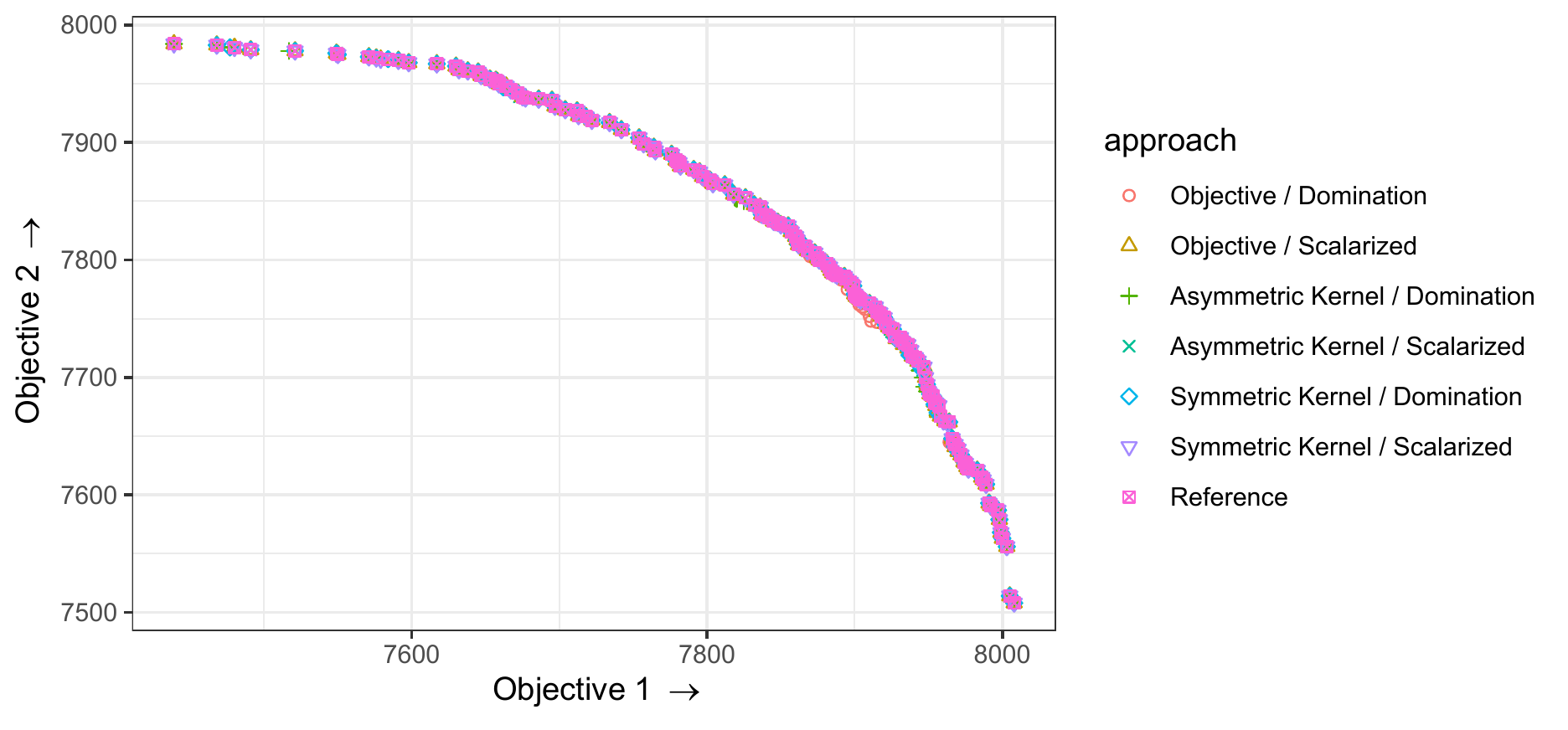}
        \subcaption{Final Fronts}
    \end{subfigure}
    \caption{MaxCut vs MaxCut $l=100$}
\end{figure}
\begin{table}
    \centering
    \caption[Best-of-Traps vs MaxCut statistical significance hypervolume]{Is x statistically significantly better than y on Best-of-Traps vs MaxCut in hypervolume at the budget limit}
    \begin{tabular}{llrrrrrr}
\toprule
{} & {approach y} & {\textbf{(O~/~D)}} & {\textbf{(O~/~S)}} & {\textbf{(AK~/~D)}} & {\textbf{(AK~/~S)}} & {\textbf{(SK~/~D)}} & {\textbf{(SK~/~S)}} \\
{fns} & {approach x} & {} & {} & {} & {} & {} & {} \\
\midrule
\multirow[c]{6}{*}{1} & \textbf{(O~/~D)} & {\cellcolor[HTML]{D3D3D3}} \color[HTML]{000000} - & {\cellcolor[HTML]{808080}} \color[HTML]{FFFFFF} 1.000 & {\cellcolor[HTML]{808080}} \color[HTML]{FFFFFF} 0.655 & {\cellcolor[HTML]{808080}} \color[HTML]{FFFFFF} 1.000 & {\cellcolor[HTML]{808080}} \color[HTML]{FFFFFF} 0.991 & {\cellcolor[HTML]{808080}} \color[HTML]{FFFFFF} 1.000 \\
 & \textbf{(O~/~S)} & {\cellcolor[HTML]{FFFFFF}} \color[HTML]{000000} 0.000 & {\cellcolor[HTML]{D3D3D3}} \color[HTML]{000000} - & {\cellcolor[HTML]{FFFFFF}} \color[HTML]{000000} 0.000 & {\cellcolor[HTML]{FFFFFF}} \color[HTML]{000000} 0.000 & {\cellcolor[HTML]{FFFFFF}} \color[HTML]{000000} 0.000 & {\cellcolor[HTML]{808080}} \color[HTML]{FFFFFF} 0.767 \\
 & \textbf{(AK~/~D)} & {\cellcolor[HTML]{808080}} \color[HTML]{FFFFFF} 0.345 & {\cellcolor[HTML]{808080}} \color[HTML]{FFFFFF} 1.000 & {\cellcolor[HTML]{D3D3D3}} \color[HTML]{000000} - & {\cellcolor[HTML]{808080}} \color[HTML]{FFFFFF} 1.000 & {\cellcolor[HTML]{808080}} \color[HTML]{FFFFFF} 0.999 & {\cellcolor[HTML]{808080}} \color[HTML]{FFFFFF} 1.000 \\
 & \textbf{(AK~/~S)} & {\cellcolor[HTML]{FFFFFF}} \color[HTML]{000000} 0.000 & {\cellcolor[HTML]{808080}} \color[HTML]{FFFFFF} 1.000 & {\cellcolor[HTML]{FFFFFF}} \color[HTML]{000000} 0.000 & {\cellcolor[HTML]{D3D3D3}} \color[HTML]{000000} - & {\cellcolor[HTML]{808080}} \color[HTML]{FFFFFF} 0.023 & {\cellcolor[HTML]{808080}} \color[HTML]{FFFFFF} 1.000 \\
 & \textbf{(SK~/~D)} & {\cellcolor[HTML]{808080}} \color[HTML]{FFFFFF} 0.009 & {\cellcolor[HTML]{808080}} \color[HTML]{FFFFFF} 1.000 & {\cellcolor[HTML]{FFFFFF}} \color[HTML]{000000} 0.001 & {\cellcolor[HTML]{808080}} \color[HTML]{FFFFFF} 0.977 & {\cellcolor[HTML]{D3D3D3}} \color[HTML]{000000} - & {\cellcolor[HTML]{808080}} \color[HTML]{FFFFFF} 1.000 \\
 & \textbf{(SK~/~S)} & {\cellcolor[HTML]{FFFFFF}} \color[HTML]{000000} 0.000 & {\cellcolor[HTML]{808080}} \color[HTML]{FFFFFF} 0.233 & {\cellcolor[HTML]{FFFFFF}} \color[HTML]{000000} 0.000 & {\cellcolor[HTML]{FFFFFF}} \color[HTML]{000000} 0.000 & {\cellcolor[HTML]{FFFFFF}} \color[HTML]{000000} 0.000 & {\cellcolor[HTML]{D3D3D3}} \color[HTML]{000000} - \\
\midrule
 \multirow[c]{6}{*}{2} & \textbf{(O~/~D)} & {\cellcolor[HTML]{D3D3D3}} \color[HTML]{000000} - & {\cellcolor[HTML]{808080}} \color[HTML]{FFFFFF} 0.951 & {\cellcolor[HTML]{808080}} \color[HTML]{FFFFFF} 0.200 & {\cellcolor[HTML]{808080}} \color[HTML]{FFFFFF} 1.000 & {\cellcolor[HTML]{808080}} \color[HTML]{FFFFFF} 0.947 & {\cellcolor[HTML]{808080}} \color[HTML]{FFFFFF} 1.000 \\
 & \textbf{(O~/~S)} & {\cellcolor[HTML]{808080}} \color[HTML]{FFFFFF} 0.049 & {\cellcolor[HTML]{D3D3D3}} \color[HTML]{000000} - & {\cellcolor[HTML]{808080}} \color[HTML]{FFFFFF} 0.019 & {\cellcolor[HTML]{808080}} \color[HTML]{FFFFFF} 0.597 & {\cellcolor[HTML]{808080}} \color[HTML]{FFFFFF} 0.141 & {\cellcolor[HTML]{808080}} \color[HTML]{FFFFFF} 0.998 \\
 & \textbf{(AK~/~D)} & {\cellcolor[HTML]{808080}} \color[HTML]{FFFFFF} 0.800 & {\cellcolor[HTML]{808080}} \color[HTML]{FFFFFF} 0.981 & {\cellcolor[HTML]{D3D3D3}} \color[HTML]{000000} - & {\cellcolor[HTML]{808080}} \color[HTML]{FFFFFF} 1.000 & {\cellcolor[HTML]{808080}} \color[HTML]{FFFFFF} 1.000 & {\cellcolor[HTML]{808080}} \color[HTML]{FFFFFF} 1.000 \\
 & \textbf{(AK~/~S)} & {\cellcolor[HTML]{FFFFFF}} \color[HTML]{000000} 0.000 & {\cellcolor[HTML]{808080}} \color[HTML]{FFFFFF} 0.403 & {\cellcolor[HTML]{FFFFFF}} \color[HTML]{000000} 0.000 & {\cellcolor[HTML]{D3D3D3}} \color[HTML]{000000} - & {\cellcolor[HTML]{808080}} \color[HTML]{FFFFFF} 0.007 & {\cellcolor[HTML]{808080}} \color[HTML]{FFFFFF} 1.000 \\
 & \textbf{(SK~/~D)} & {\cellcolor[HTML]{808080}} \color[HTML]{FFFFFF} 0.053 & {\cellcolor[HTML]{808080}} \color[HTML]{FFFFFF} 0.859 & {\cellcolor[HTML]{FFFFFF}} \color[HTML]{000000} 0.000 & {\cellcolor[HTML]{808080}} \color[HTML]{FFFFFF} 0.993 & {\cellcolor[HTML]{D3D3D3}} \color[HTML]{000000} - & {\cellcolor[HTML]{808080}} \color[HTML]{FFFFFF} 1.000 \\
 & \textbf{(SK~/~S)} & {\cellcolor[HTML]{FFFFFF}} \color[HTML]{000000} 0.000 & {\cellcolor[HTML]{FFFFFF}} \color[HTML]{000000} 0.002 & {\cellcolor[HTML]{FFFFFF}} \color[HTML]{000000} 0.000 & {\cellcolor[HTML]{FFFFFF}} \color[HTML]{000000} 0.000 & {\cellcolor[HTML]{FFFFFF}} \color[HTML]{000000} 0.000 & {\cellcolor[HTML]{D3D3D3}} \color[HTML]{000000} - \\
\midrule
 \multirow[c]{6}{*}{4} & \textbf{(O~/~D)} & {\cellcolor[HTML]{D3D3D3}} \color[HTML]{000000} - & {\cellcolor[HTML]{808080}} \color[HTML]{FFFFFF} 1.000 & {\cellcolor[HTML]{808080}} \color[HTML]{FFFFFF} 0.608 & {\cellcolor[HTML]{808080}} \color[HTML]{FFFFFF} 1.000 & {\cellcolor[HTML]{808080}} \color[HTML]{FFFFFF} 1.000 & {\cellcolor[HTML]{808080}} \color[HTML]{FFFFFF} 1.000 \\
 & \textbf{(O~/~S)} & {\cellcolor[HTML]{FFFFFF}} \color[HTML]{000000} 0.000 & {\cellcolor[HTML]{D3D3D3}} \color[HTML]{000000} - & {\cellcolor[HTML]{FFFFFF}} \color[HTML]{000000} 0.001 & {\cellcolor[HTML]{808080}} \color[HTML]{FFFFFF} 0.796 & {\cellcolor[HTML]{808080}} \color[HTML]{FFFFFF} 0.401 & {\cellcolor[HTML]{808080}} \color[HTML]{FFFFFF} 0.998 \\
 & \textbf{(AK~/~D)} & {\cellcolor[HTML]{808080}} \color[HTML]{FFFFFF} 0.392 & {\cellcolor[HTML]{808080}} \color[HTML]{FFFFFF} 0.999 & {\cellcolor[HTML]{D3D3D3}} \color[HTML]{000000} - & {\cellcolor[HTML]{808080}} \color[HTML]{FFFFFF} 1.000 & {\cellcolor[HTML]{808080}} \color[HTML]{FFFFFF} 0.999 & {\cellcolor[HTML]{808080}} \color[HTML]{FFFFFF} 1.000 \\
 & \textbf{(AK~/~S)} & {\cellcolor[HTML]{FFFFFF}} \color[HTML]{000000} 0.000 & {\cellcolor[HTML]{808080}} \color[HTML]{FFFFFF} 0.204 & {\cellcolor[HTML]{FFFFFF}} \color[HTML]{000000} 0.000 & {\cellcolor[HTML]{D3D3D3}} \color[HTML]{000000} - & {\cellcolor[HTML]{808080}} \color[HTML]{FFFFFF} 0.011 & {\cellcolor[HTML]{808080}} \color[HTML]{FFFFFF} 0.988 \\
 & \textbf{(SK~/~D)} & {\cellcolor[HTML]{FFFFFF}} \color[HTML]{000000} 0.000 & {\cellcolor[HTML]{808080}} \color[HTML]{FFFFFF} 0.599 & {\cellcolor[HTML]{FFFFFF}} \color[HTML]{000000} 0.001 & {\cellcolor[HTML]{808080}} \color[HTML]{FFFFFF} 0.989 & {\cellcolor[HTML]{D3D3D3}} \color[HTML]{000000} - & {\cellcolor[HTML]{808080}} \color[HTML]{FFFFFF} 1.000 \\
 & \textbf{(SK~/~S)} & {\cellcolor[HTML]{FFFFFF}} \color[HTML]{000000} 0.000 & {\cellcolor[HTML]{FFFFFF}} \color[HTML]{000000} 0.002 & {\cellcolor[HTML]{FFFFFF}} \color[HTML]{000000} 0.000 & {\cellcolor[HTML]{808080}} \color[HTML]{FFFFFF} 0.012 & {\cellcolor[HTML]{FFFFFF}} \color[HTML]{000000} 0.000 & {\cellcolor[HTML]{D3D3D3}} \color[HTML]{000000} - \\
\midrule
 \multirow[c]{6}{*}{8} & \textbf{(O~/~D)} & {\cellcolor[HTML]{D3D3D3}} \color[HTML]{000000} - & {\cellcolor[HTML]{808080}} \color[HTML]{FFFFFF} 0.996 & {\cellcolor[HTML]{808080}} \color[HTML]{FFFFFF} 0.998 & {\cellcolor[HTML]{808080}} \color[HTML]{FFFFFF} 1.000 & {\cellcolor[HTML]{808080}} \color[HTML]{FFFFFF} 1.000 & {\cellcolor[HTML]{808080}} \color[HTML]{FFFFFF} 1.000 \\
 & \textbf{(O~/~S)} & {\cellcolor[HTML]{FFFFFF}} \color[HTML]{000000} 0.004 & {\cellcolor[HTML]{D3D3D3}} \color[HTML]{000000} - & {\cellcolor[HTML]{808080}} \color[HTML]{FFFFFF} 0.685 & {\cellcolor[HTML]{808080}} \color[HTML]{FFFFFF} 1.000 & {\cellcolor[HTML]{808080}} \color[HTML]{FFFFFF} 0.995 & {\cellcolor[HTML]{808080}} \color[HTML]{FFFFFF} 1.000 \\
 & \textbf{(AK~/~D)} & {\cellcolor[HTML]{FFFFFF}} \color[HTML]{000000} 0.002 & {\cellcolor[HTML]{808080}} \color[HTML]{FFFFFF} 0.315 & {\cellcolor[HTML]{D3D3D3}} \color[HTML]{000000} - & {\cellcolor[HTML]{808080}} \color[HTML]{FFFFFF} 1.000 & {\cellcolor[HTML]{808080}} \color[HTML]{FFFFFF} 0.999 & {\cellcolor[HTML]{808080}} \color[HTML]{FFFFFF} 1.000 \\
 & \textbf{(AK~/~S)} & {\cellcolor[HTML]{FFFFFF}} \color[HTML]{000000} 0.000 & {\cellcolor[HTML]{FFFFFF}} \color[HTML]{000000} 0.000 & {\cellcolor[HTML]{FFFFFF}} \color[HTML]{000000} 0.000 & {\cellcolor[HTML]{D3D3D3}} \color[HTML]{000000} - & {\cellcolor[HTML]{FFFFFF}} \color[HTML]{000000} 0.000 & {\cellcolor[HTML]{808080}} \color[HTML]{FFFFFF} 0.720 \\
 & \textbf{(SK~/~D)} & {\cellcolor[HTML]{FFFFFF}} \color[HTML]{000000} 0.000 & {\cellcolor[HTML]{FFFFFF}} \color[HTML]{000000} 0.005 & {\cellcolor[HTML]{FFFFFF}} \color[HTML]{000000} 0.001 & {\cellcolor[HTML]{808080}} \color[HTML]{FFFFFF} 1.000 & {\cellcolor[HTML]{D3D3D3}} \color[HTML]{000000} - & {\cellcolor[HTML]{808080}} \color[HTML]{FFFFFF} 1.000 \\
 & \textbf{(SK~/~S)} & {\cellcolor[HTML]{FFFFFF}} \color[HTML]{000000} 0.000 & {\cellcolor[HTML]{FFFFFF}} \color[HTML]{000000} 0.000 & {\cellcolor[HTML]{FFFFFF}} \color[HTML]{000000} 0.000 & {\cellcolor[HTML]{808080}} \color[HTML]{FFFFFF} 0.280 & {\cellcolor[HTML]{FFFFFF}} \color[HTML]{000000} 0.000 & {\cellcolor[HTML]{D3D3D3}} \color[HTML]{000000} - \\
\bottomrule
\end{tabular}

\textbf{(O~/~D):}~Objective~/~Domination \hspace{2em}
\textbf{(O~/~S):}~Objective~/~Scalarized \hspace{2em}
\textbf{(AK~/~D):}~Asymmetric Kernel~/~Domination \hspace{2em}
\textbf{(AK~/~S):}~Asymmetric Kernel~/~Scalarized \hspace{2em}
\textbf{(SK~/~D):}~Symmetric Kernel~/~Domination \hspace{2em}
\textbf{(SK~/~S):}~Symmetric Kernel~/~Scalarized

\end{table}
\begin{table}
    \centering
    \caption[MaxCut vs MaxCut statistical significance hypervolume]{Is x statistically significantly better than y on MaxCut vs MaxCut in hypervolume at the budget limit}
    \begin{tabular}{lrrrrrr}
\toprule
{approach y} & {\textbf{(O~/~D)}} & {\textbf{(O~/~S)}} & {\textbf{(AK~/~D)}} & {\textbf{(AK~/~S)}} & {\textbf{(SK~/~D)}} & {\textbf{(SK~/~S)}} \\
{approach x} & {} & {} & {} & {} & {} & {} \\
\midrule
\textbf{(O~/~D)} & {\cellcolor[HTML]{D3D3D3}} \color[HTML]{000000} - & {\cellcolor[HTML]{808080}} \color[HTML]{FFFFFF} 1.000 & {\cellcolor[HTML]{808080}} \color[HTML]{FFFFFF} 0.992 & {\cellcolor[HTML]{808080}} \color[HTML]{FFFFFF} 1.000 & {\cellcolor[HTML]{808080}} \color[HTML]{FFFFFF} 1.000 & {\cellcolor[HTML]{808080}} \color[HTML]{FFFFFF} 1.000 \\
\textbf{(O~/~S)} & {\cellcolor[HTML]{FFFFFF}} \color[HTML]{000000} 0.000 & {\cellcolor[HTML]{D3D3D3}} \color[HTML]{000000} - & {\cellcolor[HTML]{FFFFFF}} \color[HTML]{000000} 0.000 & {\cellcolor[HTML]{808080}} \color[HTML]{FFFFFF} 0.910 & {\cellcolor[HTML]{FFFFFF}} \color[HTML]{000000} 0.004 & {\cellcolor[HTML]{808080}} \color[HTML]{FFFFFF} 0.987 \\
\textbf{(AK~/~D)} & {\cellcolor[HTML]{808080}} \color[HTML]{FFFFFF} 0.008 & {\cellcolor[HTML]{808080}} \color[HTML]{FFFFFF} 1.000 & {\cellcolor[HTML]{D3D3D3}} \color[HTML]{000000} - & {\cellcolor[HTML]{808080}} \color[HTML]{FFFFFF} 1.000 & {\cellcolor[HTML]{808080}} \color[HTML]{FFFFFF} 1.000 & {\cellcolor[HTML]{808080}} \color[HTML]{FFFFFF} 1.000 \\
\textbf{(AK~/~S)} & {\cellcolor[HTML]{FFFFFF}} \color[HTML]{000000} 0.000 & {\cellcolor[HTML]{808080}} \color[HTML]{FFFFFF} 0.090 & {\cellcolor[HTML]{FFFFFF}} \color[HTML]{000000} 0.000 & {\cellcolor[HTML]{D3D3D3}} \color[HTML]{000000} - & {\cellcolor[HTML]{FFFFFF}} \color[HTML]{000000} 0.000 & {\cellcolor[HTML]{808080}} \color[HTML]{FFFFFF} 0.949 \\
\textbf{(SK~/~D)} & {\cellcolor[HTML]{FFFFFF}} \color[HTML]{000000} 0.000 & {\cellcolor[HTML]{808080}} \color[HTML]{FFFFFF} 0.996 & {\cellcolor[HTML]{FFFFFF}} \color[HTML]{000000} 0.000 & {\cellcolor[HTML]{808080}} \color[HTML]{FFFFFF} 1.000 & {\cellcolor[HTML]{D3D3D3}} \color[HTML]{000000} - & {\cellcolor[HTML]{808080}} \color[HTML]{FFFFFF} 1.000 \\
\textbf{(SK~/~S)} & {\cellcolor[HTML]{FFFFFF}} \color[HTML]{000000} 0.000 & {\cellcolor[HTML]{808080}} \color[HTML]{FFFFFF} 0.013 & {\cellcolor[HTML]{FFFFFF}} \color[HTML]{000000} 0.000 & {\cellcolor[HTML]{808080}} \color[HTML]{FFFFFF} 0.051 & {\cellcolor[HTML]{FFFFFF}} \color[HTML]{000000} 0.000 & {\cellcolor[HTML]{D3D3D3}} \color[HTML]{000000} - \\
\bottomrule
\end{tabular}

\textbf{(O~/~D):}~Objective~/~Domination \hspace{2em}
\textbf{(O~/~S):}~Objective~/~Scalarized \hspace{2em}
\textbf{(AK~/~D):}~Asymmetric Kernel~/~Domination \hspace{2em}
\textbf{(AK~/~S):}~Asymmetric Kernel~/~Scalarized \hspace{2em}
\textbf{(SK~/~D):}~Symmetric Kernel~/~Domination \hspace{2em}
\textbf{(SK~/~S):}~Symmetric Kernel~/~Scalarized

\end{table}

\newgeometry{left=3cm,bottom=0.5cm} 
\begin{landscape}
    \pagestyle{empty}
\begin{table}
    \centering
    \caption{Quantiles of Hypervolume at evaluation limit for MaxCut vs Maxcut}
    \resizebox{27cm}{!}{
\begin{tabular}{rrrrrrrrrrrrrrrrrrr}
\toprule
 & \multicolumn{3}{c}{Asymmetric Kernel / Domination} & \multicolumn{3}{c}{Asymmetric Kernel / Scalarized} & \multicolumn{3}{c}{Objective / Domination} & \multicolumn{3}{c}{Objective / Scalarized} & \multicolumn{3}{c}{Symmetric Kernel / Domination} & \multicolumn{3}{c}{Symmetric Kernel / Scalarized} \\ 
 \cmidrule(lr){2-4} \cmidrule(lr){5-7} \cmidrule(lr){8-10} \cmidrule(lr){11-13} \cmidrule(lr){14-16} \cmidrule(lr){17-19}
l & 0.05 & 0.5 & 0.95 & 0.05 & 0.5 & 0.95 & 0.05 & 0.5 & 0.95 & 0.05 & 0.5 & 0.95 & 0.05 & 0.5 & 0.95 & 0.05 & 0.5 & 0.95 \\ 
\midrule
12 & 1.0000000 & 1.0000000 & 1.0000000 & 1.0000000 & 1.0000000 & 1.0000000 & 1.0000000 & 1.0000000 & 1.0000000 & 1.0000000 & 1.0000000 & 1.0000000 & 1.0000000 & 1.0000000 & 1.0000000 & 1.0000000 & 1.0000000 & 1.0000000 \\ 
25 & 1.0000000 & 1.0000000 & 1.0000000 & 1.0000000 & 1.0000000 & 1.0000000 & 1.0000000 & 1.0000000 & 1.0000000 & 1.0000000 & 1.0000000 & 1.0000000 & 1.0000000 & 1.0000000 & 1.0000000 & 1.0000000 & 1.0000000 & 1.0000000 \\ 
50 & 1.0000000 & 1.0000000 & 1.0000000 & 1.0000000 & 1.0000000 & 1.0000000 & 0.9998743 & 1.0000000 & 1.0000000 & 0.9999309 & 1.0000000 & 1.0000000 & 1.0000000 & 1.0000000 & 1.0000000 & 1.0000000 & 1.0000000 & 1.0000000 \\ 
100 & 0.9975306 & 0.9990018 & 0.9995855 & 0.9994609 & 0.9997946 & 0.9999792 & 0.9948829 & 0.9983849 & 0.9994716 & 0.9985707 & 0.9997456 & 0.9999771 & 0.9988646 & 0.9994212 & 0.9997552 & 0.9996268 & 0.9998296 & 0.9999907 \\ 
 \bottomrule
\end{tabular}
}
    
\end{table}

\begin{table}
    \centering
    \caption{Quantiles of Hypervolume at evaluation limit for Best-of-Traps vs Maxcut}
    \resizebox{27cm}{!}{
\begin{tabular}{rrrrrrrrrrrrrrrrrrr}
\toprule
 & \multicolumn{3}{c}{Asymmetric Kernel / Domination} & \multicolumn{3}{c}{Asymmetric Kernel / Scalarized} & \multicolumn{3}{c}{Objective / Domination} & \multicolumn{3}{c}{Objective / Scalarized} & \multicolumn{3}{c}{Symmetric Kernel / Domination} & \multicolumn{3}{c}{Symmetric Kernel / Scalarized} \\ 
 \cmidrule(lr){2-4} \cmidrule(lr){5-7} \cmidrule(lr){8-10} \cmidrule(lr){11-13} \cmidrule(lr){14-16} \cmidrule(lr){17-19}
l & 0.05 & 0.5 & 0.95 & 0.05 & 0.5 & 0.95 & 0.05 & 0.5 & 0.95 & 0.05 & 0.5 & 0.95 & 0.05 & 0.5 & 0.95 & 0.05 & 0.5 & 0.95 \\ 
\midrule
\multicolumn{1}{l}{fns:  1} \\ 
\midrule
12 & 1.0000000 & 1.0000000 & 1.0000000 & 1.0000000 & 1.0000000 & 1.0000000 & 1.0000000 & 1.0000000 & 1.0000000 & 1.0000000 & 1.0000000 & 1.0000000 & 1.0000000 & 1.0000000 & 1.0000000 & 1.0000000 & 1.0000000 & 1.0000000 \\ 
25 & 1.0000000 & 1.0000000 & 1.0000000 & 1.0000000 & 1.0000000 & 1.0000000 & 1.0000000 & 1.0000000 & 1.0000000 & 1.0000000 & 1.0000000 & 1.0000000 & 1.0000000 & 1.0000000 & 1.0000000 & 1.0000000 & 1.0000000 & 1.0000000 \\ 
50 & 1.0000000 & 1.0000000 & 1.0000000 & 1.0000000 & 1.0000000 & 1.0000000 & 1.0000000 & 1.0000000 & 1.0000000 & 1.0000000 & 1.0000000 & 1.0000000 & 1.0000000 & 1.0000000 & 1.0000000 & 1.0000000 & 1.0000000 & 1.0000000 \\ 
100 & 0.9913288 & 0.9994905 & 0.9996804 & 0.9994442 & 0.9996526 & 0.9997221 & 0.9881652 & 0.9993284 & 0.9997267 & 0.9996758 & 0.9996758 & 0.9998402 & 0.9918222 & 0.9996294 & 0.9997012 & 0.9996503 & 0.9997221 & 0.9998865 \\ 
\midrule
\multicolumn{1}{l}{fns:  2} \\ 
\midrule
12 & 1.0000000 & 1.0000000 & 1.0000000 & 1.0000000 & 1.0000000 & 1.0000000 & 1.0000000 & 1.0000000 & 1.0000000 & 1.0000000 & 1.0000000 & 1.0000000 & 1.0000000 & 1.0000000 & 1.0000000 & 1.0000000 & 1.0000000 & 1.0000000 \\ 
25 & 1.0000000 & 1.0000000 & 1.0000000 & 1.0000000 & 1.0000000 & 1.0000000 & 1.0000000 & 1.0000000 & 1.0000000 & 1.0000000 & 1.0000000 & 1.0000000 & 1.0000000 & 1.0000000 & 1.0000000 & 1.0000000 & 1.0000000 & 1.0000000 \\ 
50 & 0.9961656 & 0.9994964 & 1.0000000 & 1.0000000 & 1.0000000 & 1.0000000 & 0.9889824 & 0.9961694 & 1.0000000 & 1.0000000 & 1.0000000 & 1.0000000 & 0.9961732 & 1.0000000 & 1.0000000 & 1.0000000 & 1.0000000 & 1.0000000 \\ 
100 & 0.9888664 & 0.9990041 & 0.9996757 & 0.9994905 & 0.9996757 & 0.9997684 & 0.9850617 & 0.9994441 & 0.9996757 & 0.9536035 & 0.9996989 & 0.9999120 & 0.9936538 & 0.9995831 & 0.9997221 & 0.9996966 & 0.9997221 & 0.9998448 \\ 
\midrule
\multicolumn{1}{l}{fns:  4} \\ 
\midrule
12 & 1.0000000 & 1.0000000 & 1.0000000 & 1.0000000 & 1.0000000 & 1.0000000 & 1.0000000 & 1.0000000 & 1.0000000 & 1.0000000 & 1.0000000 & 1.0000000 & 1.0000000 & 1.0000000 & 1.0000000 & 1.0000000 & 1.0000000 & 1.0000000 \\ 
25 & 0.9994547 & 1.0000000 & 1.0000000 & 1.0000000 & 1.0000000 & 1.0000000 & 1.0000000 & 1.0000000 & 1.0000000 & 1.0000000 & 1.0000000 & 1.0000000 & 1.0000000 & 1.0000000 & 1.0000000 & 1.0000000 & 1.0000000 & 1.0000000 \\ 
50 & 0.9679812 & 0.9752827 & 0.9991981 & 0.9993783 & 1.0000000 & 1.0000000 & 0.9620132 & 0.9724852 & 0.9988561 & 0.9721246 & 0.9891209 & 0.9963384 & 0.9721246 & 0.9891209 & 0.9994778 & 1.0000000 & 1.0000000 & 1.0000000 \\ 
100 & 0.9438709 & 0.9861373 & 0.9986991 & 0.9895695 & 0.9980394 & 0.9988721 & 0.9096496 & 0.9930571 & 0.9968607 & 0.9596551 & 0.9963556 & 0.9991489 & 0.9879802 & 0.9965862 & 0.9987337 & 0.9967431 & 0.9981547 & 0.9995848 \\ 
\midrule
\multicolumn{1}{l}{fns:  8} \\ 
\midrule
12 & 1.0000000 & 1.0000000 & 1.0000000 & 1.0000000 & 1.0000000 & 1.0000000 & 1.0000000 & 1.0000000 & 1.0000000 & 1.0000000 & 1.0000000 & 1.0000000 & 1.0000000 & 1.0000000 & 1.0000000 & 1.0000000 & 1.0000000 & 1.0000000 \\ 
25 & 0.9985406 & 1.0000000 & 1.0000000 & 1.0000000 & 1.0000000 & 1.0000000 & 1.0000000 & 1.0000000 & 1.0000000 & 1.0000000 & 1.0000000 & 1.0000000 & 0.9985406 & 1.0000000 & 1.0000000 & 1.0000000 & 1.0000000 & 1.0000000 \\ 
50 & 0.9549993 & 0.9771493 & 0.9905488 & 0.9993854 & 1.0000000 & 1.0000000 & 0.9380138 & 0.9667504 & 0.9893503 & 0.9538636 & 0.9828897 & 0.9946051 & 0.9606475 & 0.9887283 & 0.9932474 & 0.9993854 & 1.0000000 & 1.0000000 \\ 
100 & 0.9283039 & 0.9499761 & 0.9919218 & 0.9903463 & 0.9976778 & 0.9994182 & 0.8223812 & 0.8998368 & 0.9804787 & 0.8947241 & 0.9510354 & 0.9852516 & 0.9432570 & 0.9601009 & 0.9930905 & 0.9861714 & 0.9981836 & 0.9992855 \\ 
 \bottomrule
\end{tabular}
}

\end{table}

\begin{table}
    \caption{Quantiles of Hypervolume at evaluation limit for Best-of-Traps vs Best-of-Traps}
    \resizebox{27cm}{!}{
\begin{tabular}{rrrrrrrrrrrrrrrrrrr}
\toprule
 & \multicolumn{3}{c}{Asymmetric Kernel / Domination} & \multicolumn{3}{c}{Asymmetric Kernel / Scalarized} & \multicolumn{3}{c}{Objective / Domination} & \multicolumn{3}{c}{Objective / Scalarized} & \multicolumn{3}{c}{Symmetric Kernel / Domination} & \multicolumn{3}{c}{Symmetric Kernel / Scalarized} \\ 
 \cmidrule(lr){2-4} \cmidrule(lr){5-7} \cmidrule(lr){8-10} \cmidrule(lr){11-13} \cmidrule(lr){14-16} \cmidrule(lr){17-19}
l & 0.05 & 0.5 & 0.95 & 0.05 & 0.5 & 0.95 & 0.05 & 0.5 & 0.95 & 0.05 & 0.5 & 0.95 & 0.05 & 0.5 & 0.95 & 0.05 & 0.5 & 0.95 \\ 
\midrule
\multicolumn{1}{l}{fns:  1} \\ 
\midrule
12 & 1.0000000 & 1.0000000 & 1.0000000 & 1.0000000 & 1.0000000 & 1.0000000 & 1.0000000 & 1.0000000 & 1.0000000 & 1.0000000 & 1.0000000 & 1.0000000 & 1.0000000 & 1.0000000 & 1.0000000 & 1.0000000 & 1.0000000 & 1.0000000 \\ 
25 & 1.0000000 & 1.0000000 & 1.0000000 & 1.0000000 & 1.0000000 & 1.0000000 & 1.0000000 & 1.0000000 & 1.0000000 & 1.0000000 & 1.0000000 & 1.0000000 & 1.0000000 & 1.0000000 & 1.0000000 & 1.0000000 & 1.0000000 & 1.0000000 \\ 
50 & 1.0000000 & 1.0000000 & 1.0000000 & 1.0000000 & 1.0000000 & 1.0000000 & 1.0000000 & 1.0000000 & 1.0000000 & 1.0000000 & 1.0000000 & 1.0000000 & 1.0000000 & 1.0000000 & 1.0000000 & 1.0000000 & 1.0000000 & 1.0000000 \\ 
100 & 0.9962273 & 0.9971150 & 1.0000000 & 0.9962273 & 0.9975588 & 1.0000000 & 0.9962273 & 0.9997781 & 1.0000000 & 0.9968265 & 0.9977808 & 1.0000000 & 0.9962273 & 1.0000000 & 1.0000000 & 0.9975588 & 1.0000000 & 1.0000000 \\ 
\midrule
\multicolumn{1}{l}{fns:  2} \\ 
\midrule
12 & 1.0000000 & 1.0000000 & 1.0000000 & 1.0000000 & 1.0000000 & 1.0000000 & 1.0000000 & 1.0000000 & 1.0000000 & 1.0000000 & 1.0000000 & 1.0000000 & 1.0000000 & 1.0000000 & 1.0000000 & 1.0000000 & 1.0000000 & 1.0000000 \\ 
25 & 0.9986833 & 1.0000000 & 1.0000000 & 1.0000000 & 1.0000000 & 1.0000000 & 1.0000000 & 1.0000000 & 1.0000000 & 1.0000000 & 1.0000000 & 1.0000000 & 1.0000000 & 1.0000000 & 1.0000000 & 1.0000000 & 1.0000000 & 1.0000000 \\ 
50 & 0.9863247 & 1.0000000 & 1.0000000 & 1.0000000 & 1.0000000 & 1.0000000 & 0.9586989 & 0.9925617 & 1.0000000 & 0.9950721 & 1.0000000 & 1.0000000 & 0.9891587 & 1.0000000 & 1.0000000 & 1.0000000 & 1.0000000 & 1.0000000 \\ 
100 & 0.9940624 & 0.9980371 & 1.0000000 & 0.9970557 & 1.0000000 & 1.0000000 & 0.9468775 & 0.9715386 & 1.0000000 & 0.9616099 & 1.0000000 & 1.0000000 & 0.9985115 & 1.0000000 & 1.0000000 & 1.0000000 & 1.0000000 & 1.0000000 \\ 
\midrule
\multicolumn{1}{l}{fns:  4} \\ 
\midrule
12 & 1.0000000 & 1.0000000 & 1.0000000 & 1.0000000 & 1.0000000 & 1.0000000 & 1.0000000 & 1.0000000 & 1.0000000 & 1.0000000 & 1.0000000 & 1.0000000 & 1.0000000 & 1.0000000 & 1.0000000 & 1.0000000 & 1.0000000 & 1.0000000 \\ 
25 & 0.9992133 & 1.0000000 & 1.0000000 & 1.0000000 & 1.0000000 & 1.0000000 & 0.9992133 & 1.0000000 & 1.0000000 & 1.0000000 & 1.0000000 & 1.0000000 & 0.9985696 & 1.0000000 & 1.0000000 & 1.0000000 & 1.0000000 & 1.0000000 \\ 
50 & 0.9582510 & 0.9733418 & 0.9901042 & 0.9976016 & 1.0000000 & 1.0000000 & 0.9501612 & 0.9699960 & 0.9849944 & 0.9520320 & 0.9730060 & 0.9880080 & 0.9629303 & 0.9730660 & 0.9850580 & 1.0000000 & 1.0000000 & 1.0000000 \\ 
100 & 0.9345117 & 0.9602235 & 0.9904630 & 0.9954417 & 0.9969611 & 1.0000000 & 0.7599546 & 0.8768946 & 0.9449060 & 0.8141980 & 0.8981871 & 0.9536017 & 0.9337626 & 0.9685880 & 0.9935576 & 1.0000000 & 1.0000000 & 1.0000000 \\ 
\midrule
\multicolumn{1}{l}{fns:  8} \\ 
\midrule
12 & 1.0000000 & 1.0000000 & 1.0000000 & 1.0000000 & 1.0000000 & 1.0000000 & 1.0000000 & 1.0000000 & 1.0000000 & 1.0000000 & 1.0000000 & 1.0000000 & 1.0000000 & 1.0000000 & 1.0000000 & 1.0000000 & 1.0000000 & 1.0000000 \\ 
25 & 0.9984499 & 1.0000000 & 1.0000000 & 1.0000000 & 1.0000000 & 1.0000000 & 1.0000000 & 1.0000000 & 1.0000000 & 1.0000000 & 1.0000000 & 1.0000000 & 0.9984499 & 1.0000000 & 1.0000000 & 1.0000000 & 1.0000000 & 1.0000000 \\ 
50 & 0.8948793 & 0.9279709 & 0.9727290 & 1.0000000 & 1.0000000 & 1.0000000 & 0.8824463 & 0.9205626 & 0.9718701 & 0.8803500 & 0.9383182 & 0.9698946 & 0.9176409 & 0.9524637 & 0.9867671 & 1.0000000 & 1.0000000 & 1.0000000 \\ 
100 & 0.8448364 & 0.8891752 & 0.9247271 & 0.9595769 & 0.9809773 & 0.9953038 & 0.6337114 & 0.7157183 & 0.8377873 & 0.6506068 & 0.7740703 & 0.9025823 & 0.8553535 & 0.8971707 & 0.9397734 & 0.9582393 & 0.9726549 & 0.9979491 \\ 
 \bottomrule
\end{tabular}
}

\end{table}
\end{landscape}
\restoregeometry